# A Selective Macro-learning Algorithm and its Application to the $N \times N$ Sliding-Tile Puzzle

**Lev Finkelstein** LEV@HAIFA.VNET.IBM.COM
*IBM - Haifa Research Laboratory,*
*Matam, Haifa, Israel*

**Shaul Markovitch** SHAULM@CS.TECHNION.AC.IL
*Computer Science Department,*
*Technion, Haifa 32000, Israel*

## Abstract

One of the most common mechanisms used for speeding up problem solvers is macro-learning. Macros are sequences of basic operators acquired during problem solving. Macros are used by the problem solver as if they were basic operators. The major problem that macro-learning presents is the vast number of macros that are available for acquisition. Macros increase the branching factor of the search space and can severely degrade problem-solving efficiency. To make macro learning useful, a program must be selective in acquiring and utilizing macros. This paper describes a general method for selective acquisition of macros. Solvable training problems are generated in increasing order of difficulty. The only macros acquired are those that take the problem solver out of a local minimum to a better state. The utility of the method is demonstrated in several domains, including the domain of $N \times N$ sliding-tile puzzles. After learning on small puzzles, the system is able to efficiently solve puzzles of any size.

## 1. Introduction

The goal of a learning system is to modify a performance element so as to improve its performance with respect to some given criterion. The learning system does so by using its experience to generate knowledge for use by the performance element. Most of the machine learning community is concerned with improving the *classification accuracy* of *classifiers* based on *classified examples*. A smaller part of the community is concerned with improving the *speed* of *search programs* based on *problem-solving experience*. This type of learning is commonly termed *speedup learning* (Tadepalli & Natarajan, 1996).

One of the most common methods of speedup learning is the acquisition of macro-operators (Fikes, Hart, & Nilsson, 1972; Iba, 1989; Korf, 1985; Laird, Rosenbloom, & Newell, 1986; Markovitch & Scott, 1988; Minton, 1985; Ruby & Kibler, 1992). Given the traditional definition of a search space with a set of states and a set of basic operators that connect them, a macro-operator is defined as a finite sequence of basic operators. Macro-operators are typically acquired during problem solving and are used in the same manner as basic operators. The process of acquiring macros is relatively simple. A system needs only to solve training problems and pass the search tree to the acquisition procedure, which, in turn, can add any sub-sequence of operators from the tree to its macro knowledge-base. The acquired macros, however, carry costs as well as benefits. One of the most significant costs associated with using macros is the increased branching factor of the search space. When the costs outweigh the benefits, we face a phenomenon called the *utility problem* (Minton, 1988; Gratch & DeJong, 1992; Markovitch & Scott, 1993; Mooney, 1989).

Due to the very large number of macros available for acquisition, a learning program must be selective in order to obtain a macro set with high utility. The goal of this research is to demonstrate that a simple macro-learning technique, combined with the right selection mechanisms, can lead to a speedup learning algorithm that is powerful and general, yet simple as well.

We start by defining a framework for selective macro-learning and describe the general architecture of a macro learner. The *information filtering* framework (Markovitch & Scott, 1993) is used to describe the various logical components of a macro learner. In particular, the framework emphasizes the important role of the selection mechanisms used during the learning process. We continue by describing the Micro-Hillary algorithm. To make the presentation and the experiments clearer, we present the algorithm in two stages. We first describe a simplified version of the algorithm that learns macros for a single domain at a time. We then show a generalized version that can handle a family of domains.

Micro-Hillary, like all other macro-learners, is useful for *satisficing* search. Macro-learning has a negative utility when used for *optimizing search* (Markovitch & Rosdeutscher, 1992). Even when the learned macros are optimal they are not useful for optimizing search procedures: knowing the shortest way from point $A$ to point $B$ and the shortest way from $B$ to $C$ does not tell us anything about the optimality of their concatenation.

The input for the Micro-Hillary algorithm is a domain, specified by a set of basic operators, a heuristic function (not necessarily admissible) for evaluating the merit of states, and a function for generating random goal states. The Micro-Hillary algorithm performs *learning by experimentation* (Mitchell, Utgoff, & Banerji, 1983). It generates solvable training problems with an increasing level of difficulty. The training problems are solved by a search program that performs hill-climbing combined with a procedure for escaping local minima. The escape routes are then recorded as macros. The problem solver uses the same search procedure, using macros as if they were basic operators. Micro-Hillary performs the maximal possible generalization by trying all the macros in all the states. Such a method increases the branching factor by the number of macros and would be too expensive when used in search procedures such as best-first search. Using this method in hill-climbing significantly reduces this overhead (unless the solutions found are significantly longer).

The architecture of Micro-Hillary contains building blocks that are inspired by the earlier works of many other researchers. Generating training problems with increasing difficulty was done *manually* by several researchers (for example Iba, 1985; Minton, 1985; Mitchell et al., 1983). EASe (Ruby & Kibler, 1992) used hill-climbing to avoid exponential increase in search time as a result of adding macros. Morris (Minton, 1985) and MacLearn (Iba, 1985) used macro selection methods that inspired the one used by Micro-Hillary. The Micro-Hillary architecture is a well-balanced and well-tuned combination of these building blocks that emerged after extensive experimental study.

The main domain upon which Micro-Hillary was tested is the general $N \times N$ sliding-tile puzzle (Ratner & Warmuth, 1990) which includes as special cases the 8-puzzle, 15-puzzle and the 24-puzzle. The 8-puzzle and the 15-puzzle problems have long been popular among mathematicians (Johnson & Story, 1879; Kraitchik, 1953; Tait, 1880) and AI researchers.

The sliding-tile puzzles are among the most commonly used domains for testing heuristic search algorithms and macro-learning (Bui, 1997; Iba, 1985; Korf, 1985; Laird et al., 1986; Ruby & Kibler, 1992; Schofield, 1967). The reason for the popularity of sliding-tile puzzles is the simplicity of their specification combined with the very large size of their associated search space (Pearl, 1984, p. 4). An $N \times N$ puzzle problem has an associated search graph of size $N^2!/2$. The *optimizing* $N \times N$ puzzle problem is NP-hard (Ratner & Warmuth, 1990), although some research efforts have been made in finding optimal solutions to the 15-puzzle (Korf, 1985) and the 24-puzzle (Korf & Taylor, 1996). It is not difficult to devise a domain-specific algorithm for solving the *satisficing* $N \times N$ puzzle problem (see for example Ratner & Warmuth, 1990). However, the attempts to find even non-optimal solutions for the puzzle *using weak methods* were successful only for small puzzles. A notable exception is the paper by Korf (1990), who outlined a search algorithm for solving $N \times N$ puzzles. MacLearn(Iba, 1989), for example, one of the most successful macro-learners, was demonstrated solving only a single $5 \times 5$ puzzle problem.

It is therefore a worthwhile challenge to devise a simple, domain-independent, learning algorithm that will be able to acquire the ability to efficiently solve any solvable $N \times N$ puzzle problem. The paper includes a comprehensive set of experiments that test the properties of the algorithm mostly in the $N \times N$ puzzle domain. We show that the learned macros can solve very large puzzles, and supply a formal proof that the generated macros are sufficient for efficiently solving puzzles of any size.

While the $N \times N$ puzzle is the main domain tried, the Micro-Hillary algorithm is completely domain-independent. We made efforts to build a minimal domain-independent architecture that is sufficient to solve the $N \times N$ puzzle, as well as problems in other domains, without introducing domain-dependent enhancements. We include experiments in other domains to demonstrate that the algorithm is indeed domain-independent.

Section 2 defines selective macro-learning and identifies the choices to be made when designing a macro-learning system. Section 3 describes the general Micro-Hillary[1] algorithm for learning macro-operators. Section 4 describes experiments with Micro-Hillary in various domains, mainly the 15-puzzle domain. Section 5 describes an extension of Micro-Hillary that is able to learn a family of domains that are differentiated by a single numeric parameter, and describes the application of the parameterized algorithm to the family of $N \times N$ puzzle domains. Finally, Section 6 discusses and sums up Micro-Hillary's strengths and weaknesses, in comparison to other macro-learning algorithms.

## 2. Background: Selective Macro-Learning

Let $S$ be a finite set of *states*. Let $O$ be a finite set of *operators* where each operator $o \in O$ is a function $o : S \rightarrow S \cup \{\emptyset\}$. If $o(s) = \emptyset$, we say that $o(s)$ is undefined. A *problem* is a pair of states $\langle s_i, s_g \rangle$, where $s_i \in S$ is called the *initial state* and $s_g \in S$ is called the *goal state*.[2] A

1. Micro-Hillary is a simplified version of a system called Hillary (Finkelshtein & Markovitch, 1992). The program was named after Sir Edmund Hillary. We did not notice that the name Hillary had already been used by Iba, Wogulis and Langley (1988).
2. The formalization could have used a goal predicate that returns true for states that are goal states. For simplicity we assumed a single specified goal state.

*solution* is a sequence of operators $\langle o_1, \ldots, o_k \rangle$, such that $o_k(\ldots o_1(s_i)\ldots) = s_g$. A problem $p = \langle s_i, s_g \rangle$ is *solvable* if there exists a solution for $p$ (specified by $solvable(s_i, s_g)$).

A *macro-operator* is a sequence of operators $m = \langle o_1, \ldots, o_k \rangle$. A macro-operator $m$ is applied to a state $s \in S$ by applying its basic operators in a sequence. If while applying the sequence of operators we encounter an undefined application, the application of the macro is undefined.

$$m(s) = \begin{cases} o_k(\ldots o_1(s)\ldots) & \forall i \leq k\,[o_i(\ldots o_1(s)\ldots) \neq \emptyset] \\ \emptyset & \text{otherwise.} \end{cases}$$

A *problem solver* $\varphi$ is an algorithm that takes a problem $p$ and a set of operators $O$ and applies operators to states, searching for a solution to $p$. Given a solvable problem $p$, a problem solver $\varphi$ and a set of operators $O$, we define $SearchCost(\varphi, p, O)$, the *cost* of solving $p$ using $\varphi$ and $O$, as the number of operator applications performed by $\varphi$ until a solution is found. Note that SearchCost is different from the solution cost—the cost of the solution path according to some cost function. In this work we are only interested in *satisficing* search—minimizing the cost of the search regardless of the cost of the resulting solution.

The *utility* of a set of macros $M$, with respect to a problem $p$, a problem solver $\varphi$ and a set of operators $O$, is defined as

$$U_{p,\varphi,O}(M) = SearchCost(\varphi, p, O) - SearchCost(\varphi, p, O \cup M). \tag{1}$$

Thus, the utility of a set of macro operators with respect to a given problem and a given problem solver is the time saved by using these operators. When the problems are drawn from some fixed distribution $D$, we use expectation values for a problem randomly drawn from $D$:

$$U_{\varphi,O}(M) = E_{D(p)}[SearchCost(\varphi, p, O) - SearchCost(\varphi, p, O \cup M)]. \tag{2}$$

In general, the utility of knowledge depends on the criteria used for evaluating the performance of the problem solver (Markovitch & Scott, 1993). Equation 2 assumes a *speedup learner*, i.e., a learner whose goal is to increase the speed of solving problems.

Most of the macro-learning systems perform *learning by experimentation* (Mitchell et al., 1983). The program solves training problems and acquires sequences of operators applied during the search. Given a set of operators $O$, a distribution of problems $D$, and a problem solver $\varphi$, the goal of a macro-learning system is to acquire a set of macro-operators $M$, such that $U_{\varphi,O}(M)$ is positive, and as high as possible. It is quite possible, however, that the utility of $M$ will be negative, as the added macros also increase the branching factor.

Markovitch and Scott (1993) introduced the *information filtering* framework, which offers a systematic way of dealing with the utility problem by being more selective. The framework identifies five logical types of selection processes in learning systems: selective experience, selective attention, selective acquisition, selective retention and selective utilization. The framework views learning programs as information processing systems where information flows from the experience space through some attention mechanism, through an acquisition procedure to the knowledge base and finally to the problem solver. The five filters are defined with respect to their logical location within the information flow.

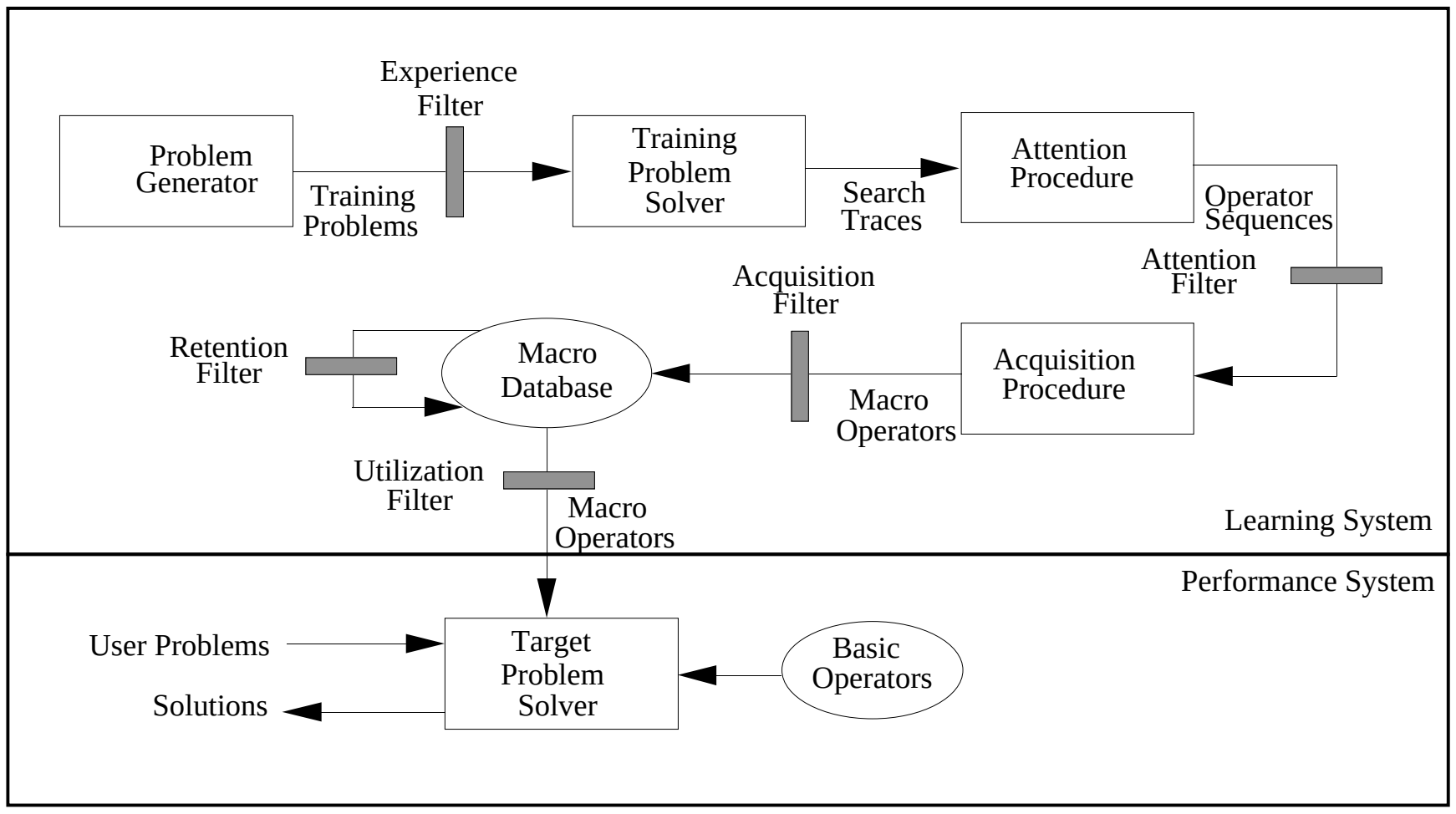

Figure 1: The architecture of an off-line macro-learning system. Training problems are generated and filtered by the experience filter. The attention procedure converts the search traces into operator sequences that are filtered by the attention filter. The sequences are converted to macro-operators. The acquisition filter decides whether to add new macros to the macro set. The retention filter deletes macros with negative utility. The utilization filter allows only a subset of the macro set to be used by the problem solver.

The general architecture offered by the information-filtering framework, instantiated to off-line macro-learning, is shown in Figure 1. Markovitch and Scott argue that the art of building successful learning systems often lies in selecting the right combination of filters. A careful examination of existing macro-learning systems reveals that those containing sophisticated filtering mechanisms performed the most successful learning.

## 3. Micro-Hillary

In this section we discuss Micro-Hillary—a particular instantiation of the architecture described above that is both very simple, yet powerful enough to perform efficient learning in a large class of problem domains. Micro-Hillary, like most macro-learners, works with *satisficing* (Simon & Kadane, 1975) search programs, i.e., programs whose goal is to find solutions as fast as possible regardless of the length of the solution found (Pearl, 1984, p. 14). Such programs typically use heuristic functions which serve as preference predicates—the search strategy prefers to expand states with a lower value. In this work we consider heuristic functions which are not necessarily underestimating, but are always positive, except in the goal states where they have a zero value. We call such heuristic functions *well-behaved*. Many of these heuristic functions order states reasonably well except in a small number of local minima. The basic idea of Micro-Hillary is to acquire macro-operators for escaping from local minima, i.e., macros that lead from a local minimum to a state with a better heuristic value.

Micro-Hillary works by generating training problems and solving them. A good way to detect local minima is to use a search method such as hill-climbing, which is susceptible to local minima. Therefore, Micro-Hillary uses simple hill-climbing[3] for its problem solver, both in the learning and in the problem-solving mode. When trapped in a local minimum, Micro-Hillary searches for a way out to a state with a better heuristic value, and records this escape route as a macro-operator. Let $Q$ be a predefined limit. We say that Micro-Hillary is in *quiescence* when it solves $Q$ consecutive training problems without acquiring any new macros. Micro-Hillary quits when it detects quiescence.

When used for solving externally supplied problems, Micro-Hillary uses the union of the basic operators and the learned macros. When faced with a local minimum, it searches for an escape route but does not acquire macros. An alternative version can perform on-line learning. Figure 2 shows the Micro-Hillary algorithm. To complete the algorithm we need to specify a method for generating training problems and a method for escaping from local minima. The *SolveProblem* procedure serves both for learning and for solving externally supplied problems.

### 3.1 Selective Experience: Generating Solvable Problems with Increasing Levels of Difficulty

The basic architecture of a macro-learning system requires a mechanism for generating training problems and a method for filtering them (or, alternatively, a method for ordering them). The main goal of the filter is to save learning resources by concentrating on helpful problems. To focus the attention of the learner, our acquisition method considers only macros that are parts of the solution path. Therefore, to save learning resources, a good problem generator should generate only solvable problems. In addition, it is desirable to order the training problems in increasing levels of difficulty. This saves learning resources by acquiring as much knowledge as possible using easy problems.

One good way to implement such a problem generation methodology is to require a domain-specific generator for each domain. For example, solvable problems in the $N \times N$ puzzle domain can be generated by performing an even permutation on the goal state. Since one of our main goals was to make the algorithm as domain-independent as possible, we have supplied a default domain-independent generator that works in many domains, including the sliding-tile puzzle. Our default problem generator assumes the following:

- The specification of the problem domain includes an algorithm for generating random goal states (unless there is only one goal state).
- All the operators are *reversible*. If there is an operator connecting state $s_1$ to state $s_2$ then there is also a sequence of operators connecting $s_2$ to $s_1$:

$$\forall s_1, s_2 \in S\left[\exists o \in O\left[s_2 = o(s_1)\right] \rightarrow \exists O' = \langle o_1, \ldots, o_k \rangle, o_i \in O\left[s_1 = O'(s_2)\right]\right].$$

  Alternatively, the specification of the domain includes a list of reverse operators.

Problems are generated by applying a random sequence of operators to a randomly generated goal state. This process is known as *random walk*. The length of the random sequence is increased after each problem generation. In some domains, when the random

3. Simple hill-climbing means that the problem solver steps forward as soon as an improvement is detected without waiting for the *best* improvement.

```
procedure MICRO-HILLARY(BasicOps,h)
   Macros ← {}
   q ← 0                                                  ; q is the quiescence counter
   loop until q > Q                                       ; Q is the quiescence parameter
      ⟨s_i, s_g⟩ ← GenerateTrainingProblem()
      q ← q + 1
      MacrosBefore ← |Macros|
      SolveProblem(s_i, s_g, BasicOps, Macros, h)
      If |Macros| > MacrosBefore then q ← 0               ; New macros were learned
   Return Macros

procedure SolveProblem(s_i, s_g, BasicOps, Macros, h)
   Solution ← {}
   s ← s_i
   loop until s = s_g
      LocalMinimum ← TRUE
      loop for o ∈ BasicOps ∪ Macros while LocalMinimum
         if h(o(s), s_g) < h(s, s_g) then          ; Progress was achieved. Step forward.
            Solution ← Solution||o
            s ← o(s)
            LocalMinimum ← FALSE
      if LocalMinimum then                                ; No progress—local minimum.
         O ← FindEscapeRoute(s, s_g)
         s ← O(s)
         Solution ← Solution||O
         if LearningMode then Macros ← Macros ∪{O}
   Return Solution
```

FindEscapeRoute$(s, s_g)$: A procedure that finds a sequence of operators $O = o_1, \ldots, o_k$ such that $h(o_k(\ldots o_1(s)\ldots), s_g) < h(s, s_g)$.

Figure 2: The MICRO-HILLARY algorithm

sequences are short enough compared to the size of the search space, and when there are no "bottlenecks" in the search graph, longer sequences are likely to yield "harder" problems (problems with larger distance between the start and goal states). In other domains the hardness of the problems may be uniformly distributed and not correlate with the length of the random walk. In such domains we can use the heuristic function as an estimate of the problem difficulty by which the random problems will be ordered. The generator sets a threshold on the heuristic distance between the initial state and the goal state and performs a random walk until a state with the required heuristic distance is found. The threshold is incremented with each iteration. This approach is still problematic in domains with "bottlenecks" where parts of the search graph are difficult to access. For such domains it is best to use a domain-dependent generator.

In Section 5 we describe a method for increasing the difficulty of training problems by increasing the parameter that characterizes the domain (such as $N$ in the case of $N \times N$ puzzles). For most parameterized domains, such as the $N \times N$ puzzle, increasing the parameter indeed increases the difficulty of problems. The framework of processing input examples by order of difficulty, called *learning from exercises*, was studied formally by Natarajan (1989) and empirically by Reddy and Tadepalli (1997). Both works assume a teacher who supplies a sequence of problems in increasing order of difficulty. Our default generator attempts to produce such a sequence automatically, freeing the learner from the need of a teacher.

### 3.2 Selective Attention: Acquiring Macros Leading from Local Minima to Better States

Given a search trace, every subpath of it can be recorded as a macro. Obviously, there are too many subpaths, so we need to employ a filter in order to acquire a set of macros with high utility. What properties of macros are likely to indicate high utility? There are three requirements that should be fulfilled:

1. The macros should be useful, i.e., they should save the system significant search resources.
2. There should be as few macros as possible. Redundant macros increase the branching factor of the search graph without associated benefits.
3. The macros should be as short as possible, for two reasons:
    - Long macros tend to be less general. The main reason is that applying a long sequence of basic operators to a state is likely to encounter an illegal state.
    - For the learning method of Micro-Hillary, longer macros require significantly more learning resources (for the escape procedure).

The filtering method we use for Micro-Hillary only acquires macros leading from a local minimum to a better state. Whenever not in a local minimum, the search procedure can use the heuristic function, so macros that do not start in a local minimum are not necessary. The macros acquired by this filtering method are useful since they enable the search procedure to continue from a local minimum. These macros are also the shortest that satisfy that condition. We call this method a *minimum-to-better* filtering. Let $h$ be a *well-behaved* heuristic function (not necessarily admissible), let $e = \langle s_{init}, s_g \rangle$ be a training problem, and let $T = \langle o_1, \ldots, o_n \rangle$ be a solution to $e$. Let $s_l = o_l(o_{l-1}(\ldots o_1(s_{init})))$ denote the state resulting from the application of the sequence $\langle o_1, \ldots, o_l \rangle$ to the initial state. A subpath $\langle o_j, o_{j+1}, \ldots, o_{j+k} \rangle$ is selected by the *minimum-to-better* attention filter if and only if the following condition holds:

$$\begin{array}{l} \forall o \in O[h(s_j, s_g) \leq h(o(s_j), s_g)] \wedge \\ h(s_{j+k}, s_g) < h(s_j, s_g) \wedge \\ \forall y[0 < y < k \rightarrow h(s_j, s_g) \leq h(s_{j+y}, s_g)] \end{array} \quad (3)$$

The first conjunct says that the first state of a macro must be a local minimum. It serves to satisfy requirement 2 above—that there should be as few macros as possible. The second conjunct says that the last state must be a better state. It serves to satisfy requirement 1 above—that macros should be useful. The third conjunct says that the macro is the shortest satisfying the former two conjuncts. It serves to satisfy the third requirement above—that macros should be as short as possible. When a sufficient number of such macros has been acquired, the search program may be able to advance toward the goal state without ever being trapped in local minima.

**Definition 1** *Let $S$ be a set of states and $O$ a set of operators. Let $s_g$ be a goal state. Let $h$ be a* well-behaved *heuristic function. Let $M$ be a set of macro-operators. We say that $M$ is* complete *with respect to $h$ and $s_g$ if and only if*

$$\forall s \in S - \{s_g\}[solvable(s, s_g) \rightarrow \exists o \in O \cup M[h(o(s), s_g) < h(s, s_g)]].$$

A *complete* macro set "smooths" the heuristic function by eliminating local minima. The length of the individual macros in a complete set depends on the distance between local minima and their associated better state.

**Definition 2** *Let $S$ be a set of states and $O$ be a set of operators. Let $s_g$ be a goal state. Let $h$ be a* well-behaved *heuristic function. Let $d(s_1, s_2)$ denote the graph distance between $s_1$ and $s_2$ in the graph defined by $O$. The* radius *of a state $s_0 \in S$ with respect to $h$, $s_g$ and $O$ is defined as:*

$$radius(s_0) = \min_{s \in S \wedge h(s,s_g) < h(s_0,s_g)} d(s_0, s).$$

*The radius of the goal state is 0. The radius of a state which is not a local minimum is 1.*

The *minimum-to-better* filter looks similar to the *minimum-to-minimum*[4] filtering method used by Iba (1989). However, there are several problems with the *minimum-to-minimum* method that the *minimum-to-better* filter avoids. When the distance between minima is large, the minimum-to-minimum filter will acquire very long macros. This violates the third requirement above. Figure 3(a) illustrates this problem. The X axis stands for the sequence of states along the solution path. The Y axis stands for the heuristic values of states. Long macros such as those acquired by the minimum-to-minimum filter tend to be much less general (they are more likely to fail). This problem is intensified with incremental learning. In the first stages of learning, the minima are close together and short macros are learned. These macros, however, make minima more sparse in later stages of learning, leading to the acquisition of very long macros. This problem becomes extreme when there is only one minimum in the solution path. In such a case, the minimum-to-minimum filter will either learn nothing, or will use Iba's approach and consider the start and end states of the solution path as minima, leading to the acquisition of two very long macros. Indeed, when we tried to replace the *minimum-to-better* with the *minimum-to-minimum* filter, Micro-Hillary went into a very long session of acquiring very long macros. Iba had to introduce an acquisition filter that blocks long macros in order to reduce the harmfulness of this phenomenon.

4. The original name given by Iba is peak-to-peak. To make the name of the filter more appropriate for heuristic functions that consider lower values as better, we call Iba's method minimum-to-minimum.

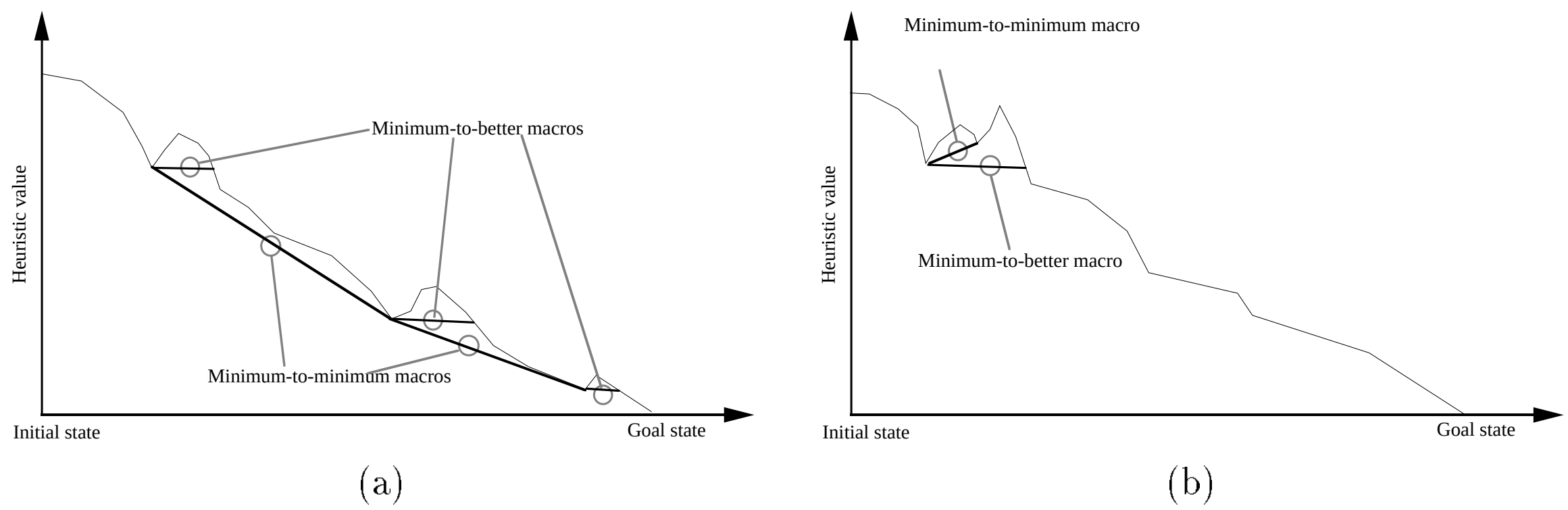

Figure 3: An illustration of the differences between the *minimum-to-minimum* and *minimum-to-better* filtering methods. The X axis stands for the sequence of states along the solution path. The Y axis stands for the heuristic values of states. (a) A situation where the *minimum-to-better* strategy acquires short macros that are just sufficient to allow progress in the heuristic search while the *minimum-to-minimum* strategy acquires unnecessarily long macros that are likely to be less applicable. (b) A situation where the *minimum-to-minimum* strategy acquires a macro that leads from a local minimum to a state with *worse* (higher) value. The *minimum-to-better* strategy always acquires macros that lead to a state with a better (lower) value.

Another problem with the *minimum-to-minimum* method is that it may learn routes from a minimum to a *worse* (higher) minimum. This violates the first requirement above. Figure 3(b) illustrates this problem. Of course, it may then learn another macro from the worse minimum, but the total number of macros still increases needlessly. The *minimum-to-better* strategy always acquires macros that lead to a state with a better (lower) value. In Section 6 we discuss further differences between MacLearn and Micro-Hillary.

The T-Macros acquired by Morris (Minton, 1985) also resemble the macros acquired by Micro-Hillary. A T-Macro is "locally anomalous. Its initial segment appears to make no progress, but the sequence as a whole is rated as advantageous" (Minton, 1985). It is therefore possible to call the selection strategy employed by Morris *any-to-better*. This method, which does not insist on starting a macro from a local minimum, will acquire more macros than the minimum-to-better strategy. These macros are not necessary and violate the second requirement above. The larger number of (possibly unnecessary) macros will increase the branching factor and may deteriorate the performance of the program that uses them.

Figure 4 illustrates the difference between the two strategies. While the *minimum-to-better* strategy will acquire one macro that smooths the path, the *any-to-better* strategy acquires many macros that will unnecessarily increase the branching factor.

### 3.3 Escaping from Local Minima

In Section 3.2 we defined the *radius* of a state as the shortest graph distance to a state with a better (lower) heuristic value. The radius affects the resources needed to find an escape route from a local minimum. There are several possible ways to escape from a local minimum.

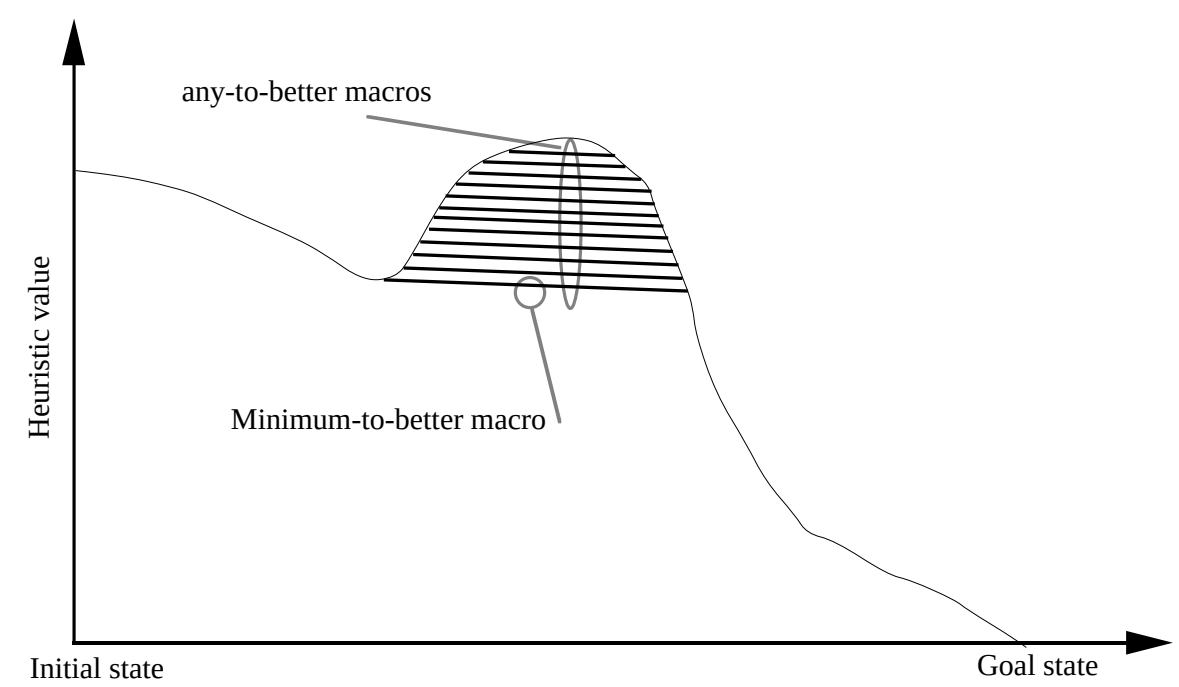

Figure 4: An illustration of the difference between the *any-to-better* and *minimum-to-better* selection methods. The X axis stands for the sequence of states along the solution path. The Y axis stands for the heuristic value of states. The any-to-better strategy acquires more (possibly unnecessary) macros.

Exhaustive search methods, such as breadth-first search (BFS) or iterative deepening (ID), guarantee finding an escape route if such a route exists. However, an exhaustive search requires too much computation for any non-trivial escape task. We can try to make some use of the heuristic function by calling A* or IDA*. Both methods, however, waste much effort to ensure that the route to the goal state is optimal. We are looking for a search method that will find solutions quickly if the heuristic function is sufficiently accurate, but also guarantees finding a solution even if the heuristic function is poor.

We came up with a search method that we call Iterative Limited BFS (ILB). The basic procedure is Limited BFS, which gets an initial state, a goal predicate, a breadth limit $B$ and a depth limit[5] $D$. It performs a BFS search to depth $D$. Whenever the number of nodes kept in memory exceeds $B$, it deletes the node with the worst heuristic value. This search is different from beam search. Limited BFS keeps the $B$ best nodes of the current depth, while beam search keeps the $B$ best nodes regardless of their depth. When $B$ approaches infinity, beam search becomes best-first search, while limited BFS becomes BFS. ILB calls limited BFS iteratively until either a full BFS to depth $D$ is performed or a solution is found. The breadth limit $B$ is increased for each iteration and is set to the value $k + b^i$, where $k$ is some constant, $b$ is the maximum branching factor and $i$ is the iteration number. This scheme guarantees that ILB will perform full BFS to depth $D$ after $D$ iterations. Figure 5 shows the algorithm. Note that this method is somewhat similar to iterative broadening (Ginsberg & Harvey, 1992), but there the broadening is performed at the node level. That means that for domains such as the sliding-tile puzzles which have a branching factor of less than 3, only 3 iterations would be performed.

Figure 6 illustrates the way that ILB works. The state at the top is the local minimum. The state at the left bottom is a state with a lower (better) heuristic value. The states are ordered in levels according to their distance from the local minimum. In each level they are ordered by their heuristic value (the lowest value to the left). The gray states are the escape

5. $D$ is a predefined upper limit ($D = 100$ was used for the experiments described here) on the maximal radius. If, for some reason, the escape route is longer than $D$, we can iterate the algorithm with a larger $D$.

```
procedure ILB(state, goal, h, D)
    b ← number of basic operators
    loop for exp = 1 to D
        BreadthLimit ← k + b^exp
        Path ← LBFS(BreadthLimit, D, State, Goal, h)
        If Path is not empty return path

procedure LBFS(BreadthLimit, State, Goal, h, D)
    InitValue ← h(State,Goal)
    NewOpen ← (Node(State,NIL))
    loop for d = 1 to D
        Open ← NewOpen; NewOpen ← NIL
        loop for Node ∈ Open
            loop for Op in Operators
                NewState ← Op(State)
                if h(NewState) < InitValue
                    return Node.Op || op
                else
                    NewOpen ← Insert(Node(NewState,Node.Op ||op), NewOpen)
                    if length(NewOpen) > BreadthLimit
                        Remove from NewOpen a node with maximum h
```

Figure 5: The ILB algorithm

route. ILB would find the route when using a breadth limit of 4 and will require linear time ($4 \times 11$) for its last iteration. ID will require exponential time for its last iteration ($3^{11}$ if the branching factor is 3). The last iteration is dominant for both algorithms. Note that ILB works differently from iterative beam search. Beam search could progress to the right portion of the search graph instead of going deeper as ILB does.

The particular escape method is not an inherent part of Micro-Hillary. In the experimental section we will show that for the $N \times N$ puzzle domain, ILB is much faster than iterative deepening. In domains where the heuristic function does not indicate the direction of the escape route, ILB may reduce to full BFS and require too much memory. Furthermore, if $D \gg Length(Escape)$, ILB will also require much more computation time because all the iterations except the last one will be performed to depth $D$, while ID would search to a depth that is equal to the length of the escape route. In such cases, using iterative deepening as an escape algorithm may be more appropriate.

One important issue to consider when designing an escape procedure is whether acquired macros are used during the escape search. Using macros increases the branching factor, which can lead to a prohibitively expensive search, but it also allows deeper penetrations. In the experiments performed, we found that it is more beneficial to search for an escape route using only basic operators. It is quite possible, however, that in some domains it is beneficial to use macros.

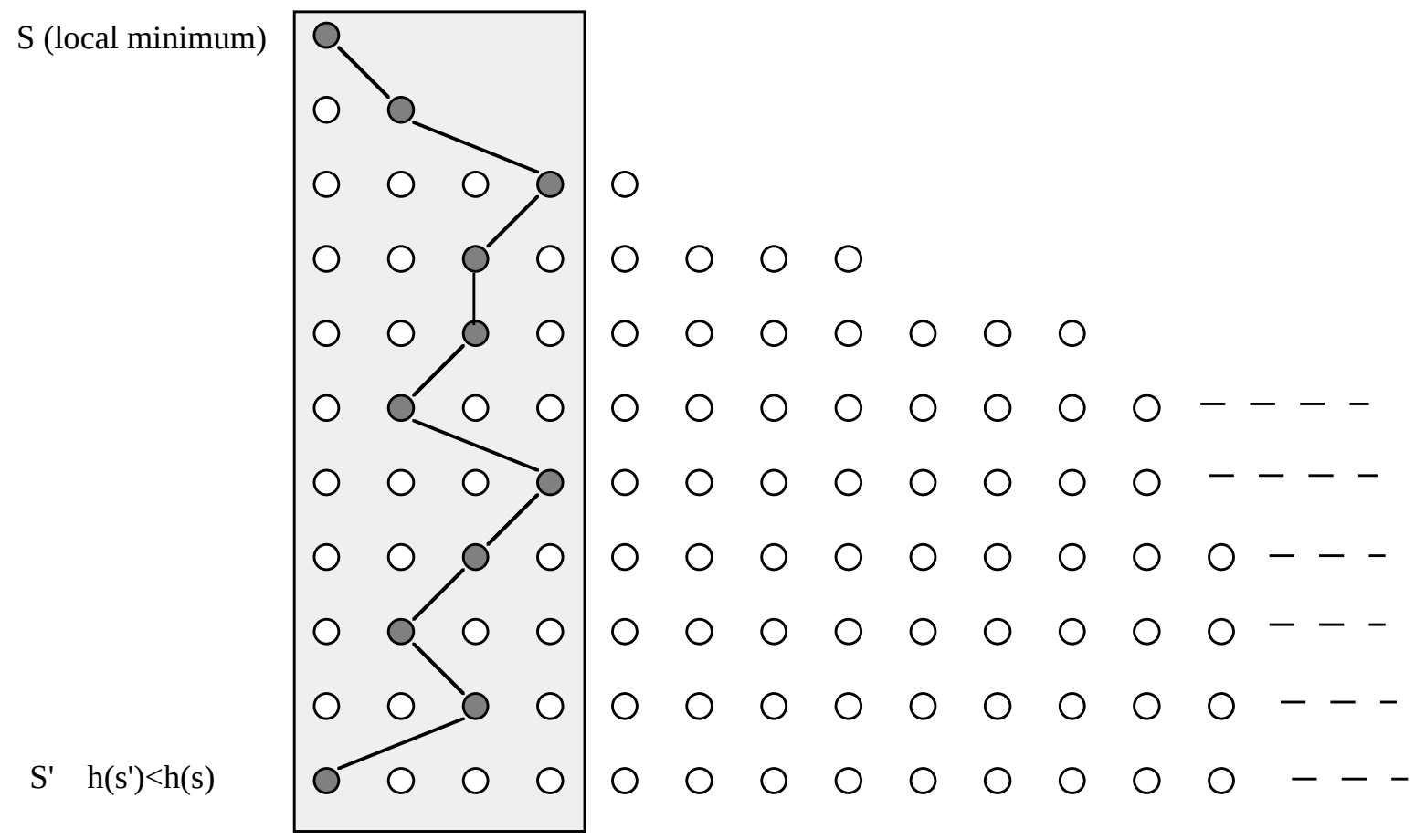

Figure 6: An illustration of the way that ILB works. The state at the top is the local minimum. The state at the bottom left is a state with a lower (better) heuristic value. The states are ordered in levels according to their distance from the local minimum. In each level they are ordered by their heuristic value (the lowest value to the left). The gray states are the escape route. The gray rectangle shows the states visited in the last iteration of ILB. ILB will find the route when using a breadth limit of 4 and will require linear time ($4 \times 11$) for its last iteration. ID will require exponential time for its last iteration ($3^{11}$ if the branching factor is 3).

### 3.4 Generalization and Use of Macros

A major factor that determines the utility of macros is the decision regarding where to use them. The major contributor to the utility problem in macro learning is the increased branching factor of the search graph as a result of the added macros. At one extreme, we can avoid generalization and apply a macro only at the same state where it was acquired. This approach will reduce the cost of the added branching factor but will also reduce the benefit of macros since they will be rarely applicable. At the other extreme, we can aim for full generalization and apply a macro at any state as if it were a basic operator. This approach has the potential of significantly increasing the benefit of macros, but it also increases the cost dramatically because every acquired macro increments the branching factor by one. MACLEARN, which uses best-first search, cannot use the second approach because it will make the search prohibitively expensive. Therefore, it uses a domain-specific pattern language and domain-specific abstraction procedures to perform a more restricted generalization of macros. Since in MICRO-HILLARY we tried to avoid domain-specific procedures, we took the second approach of full generalization. Fortunately, the hill-climbing search employed by the problem solver in MICRO-HILLARY does not suffer from the extreme increase in search time as a result of the increase in the branching factor.

### 3.5 Using Micro-Hillary for a Specific Domain

To use Micro-Hillary for learning in a specific domain, the following domain-specific information should be supplied:

1. A function that generates a random goal state.
2. A set of operators that accept a state and return either another state or $\emptyset$.
3. A well-behaved heuristic function.

### 3.6 Correctness of Micro-Hillary

A problem solver is *complete* if it terminates with a solution when one exists (Pearl, 1984, p. 75). If, when starting from a solvable state, we never reach unsolvable states, and if the range of the heuristic function is finite, then a complete set of macros yields a complete problem solver.

Let $S$ be a set of states and $O$ a set of basic operators. Let $P \subseteq S \times S$ be a set of solvable problems with the following property:

$$\forall \langle s_i, s_g \rangle \in P, \forall s \in S \left[ solvable(s_i, s) \rightarrow solvable(s, s_g) \right].$$

Let $h$ be a well-behaved heuristic function with a finite range, $R_h$, over S ($R_h = |\{h(s)|s \in S\}| < \infty$). Let $M$ be a complete set of macros. Then, SolveProblem using $M$ is a complete problem solver with respect to $P$. Furthermore, the number of operator applications is bounded by $(|O| + B_m)R_h$ where $B_m$ is the total length of macros in $M$.

**Proof:**

In every step of the hill-climbing procedure we are guaranteed (by the completeness of $M$) to find a state with a heuristic value which is lower than the value of the current state. Since $h$ is well-behaved, and since it has a finite number of values, $R_h$, we can make at most $R_h$ steps before reaching a goal state with $h(s) = 0$. At every step, in the worst case, we try all the basic operators and all the macros before finding the one that makes progress. Therefore the total number of operator applications is bounded by $(|O| + B_m)R_h$. □

## 4. Experimenting with Micro-Hillary

To test the effectiveness of the Micro-Hillary algorithm, we experimented with it in various domains. Most of the experiments were done in the 15-puzzle domain. The basic operators in this domain are *Up*, *Down*, *Left* and *Right*, indicating the direction in which the empty tile "moves". The heuristic function, which we denote by RR, is similar to the one used by Iba (1989) and by Korf (1990). The function contains three factors that are ordered lexicographically: the total number of tiles minus the count of tiles consecutively placed in a left-to-right, top-to-bottom, row-by-row order, the Manhattan distance of the first tile that is not in place to its destination, and the Manhattan distance of that tile from the empty square. Consider, for example, the following puzzle:

| | | | | |
|---|---|---|---|---|
| 1 | 2 | 3 | 4 | 5 |
| 6 | 7 | 8 | 19 | 16 |
| 14 | | 17 | 18 | 15 |
| 20 | 24 | 13 | 22 | 21 |
| 23 | 9 | 10 | 11 | 12 |

. There are 8 tiles placed in the specified order; therefore, the first component is $24-8=16$. The first tile not in its goal location is 9, so the second component (the Manhattan distance between 9 and its destination) is 5. The third component (the Manhattan distance between 9 and the empty square) is 2. This function essentially tells the problem solver to try ordering the tiles row by row.[6] This heuristic encodes subgoaling of puzzle problems which is quite natural, and is immediately inferred by most people trying to solve the sliding-tile puzzle. However, many people have difficulties solving the puzzle even after inferring this heuristic. We will also report results for other heuristic functions such as the sum of Manhattan distances.

We begin by describing the experimental methodology used, and continue with a description of the experiments performed.

### 4.1 Experimental Methodology

Most of the experiments described here consist of a learning phase and a testing phase. The training problems are generated by MICRO-HILLARY. We let every learning phase run until MICRO-HILLARY reaches quiescence and halts. Since the training problems are generated by random sequences, we repeat each learning session 100 times. We test each resulting macro set by allowing the problem solver to use the macros for solving a set of 100 test problems. A random test-problem is generated in the same manner as a training problem. The only difference is the length of the random sequence of operators applied to the goal state. For testing we used random sequences of length 1,000,000 to ensure that the problems will be sufficiently difficult (or sufficiently random). For the sliding-tile domains, the *test* problems were generated using a known domain-dependent method for creating random solvable problems. The generator continuously creates random permutations of the goal state and returns the first even permutation. The domain-specific method does not harm the generality of MICRO-HILLARY since it is used only by the experimenter. This method allows us to compare the results obtained by MICRO-HILLARY to the results obtained by other researchers. When in testing mode, we allow MICRO-HILLARY to escape from local minima but we do not allow it to acquire new macros (see the SolveProblem procedure in Figure 2). For all the experiments described in this paper, MICRO-HILLARY was able to solve all the problems while in testing mode; therefore, no special handling of "censored data" (Etzioni & Etzioni, 1994; Segre, Elkan, & Russell, 1991) was necessary. Recall that the quiescence test lets MICRO-HILLARY stop only after it is able to solve 50 problems without getting stuck in any local minima. This significantly reduces the likelihood that it will encounter a local minimum after learning.

#### 4.1.1 DEPENDENT VARIABLES

We measure three aspects of the learning process: the resources consumed during learning, the characteristics of the resulting macro set and its utility:

- *Learning resources:* The resources consumed during the learning process until the learning system decides that it has learned enough. Note that the quiescence parameter significantly affects the learning resources. A higher value for this parameter

6. Table 19 specifies the exact definition of this function for the general $N \times N$ case.

increases resource consumption, but also decreases the likelihood that the problem solver will encounter a local minimum during testing.

- *Training problems:* The total number of problems solved during training. The problem with this dependent variable is that it ignores the time invested in the search and the time invested in generating training problems.
- *Operator applications:* Since the basic operation that is used both in search and in problem generation is the application of a basic operator to a state, we use the total number of operator applications as the principle measurement for the learning resources consumed.

• *Macro-set statistics:* Statistics about the characteristics of the generated macro set.

- *Total number:* The total number of macros acquired during the learning session.
- *Mean Length:* The average length of a macro.
- *Max length:* The maximum length of a macro. This variable approximates the maximum *radius*.

• *The utility of the acquired macros:* According to Equation 2, the utility of the acquired macros depends on the cost of the search when using them. Therefore, our principle dependent variable should measure problem-solving speed.

- *CPU time:* The most obvious candidate for measuring problem-solving speed is CPU time spent during search. However, such a measurement is overly affected by irrelevant factors such as hardware[7], software and programming quality.
- *Expanded nodes:* The number of nodes expanded during the search. This is a common method for measuring search speed. Nevertheless, this measurement may be misleading in the context of macro-learning, since the branching factor increases when acquiring macros.
- *Generated nodes:* The number of nodes generated during search. This measurement takes into account the increased branching factor, but it does not account for the higher cost of generating a node by a macro due to the application of several basic operators.
- *Operator applications:* The number of applications of a basic operator to a state. Note that we count every application, including those which are part of macro-operators and those which fail. This is the principle dependent variable, as it represents most accurately the problem-solving speed.
- *Solution quality:* Macro-learning is not a suitable technique when the macros are used by an optimizing search procedure (Markovitch & Rosdeutscher, 1992). We are still interested, however, in the quality of the solution obtained. We measure the quality of the solution by its length in basic operators.

7. For example, all the CPU time figures reported here were achieved running a Common Lisp program on an outdated Sun. Recent experiments on an UltraSparc yield times which are about 4.4 times faster than those reported here.

#### 4.1.2 Independent Variables

There are many independent variables that affect the performance of the learning system. For the experiments described here we look at the following variables:

- *Selection method:*
  - *Minimum-to-better*: The main selection method defined by Equation 3.
  - *Minimum-to-minimum*: Our implementation of MacLearn's selection method.
  - *Any-to-better*: Our implementation of Morris's selection method.
- *Heuristic function:*
  - RR: The heuristic function described above.
  - *Row-by-row-2* : The RR heuristic without the third component.
  - *Reduction* : A variation of the RR heuristic that leads to placing of tiles in the first row, then the last column, then the second row, then the second to last column etc. This heuristic was suggested by Korf (1990).
  - *Spiral*: A variation of the RR heuristic that leads to placing the tiles in a spiral order from outside in.
  - *Sum-of-Manhattan-distances* : The sum of the Manhattan distances between each tile and its destination (will be denoted by MD).
- *Domain:* Most of the experiments described here were performed in the 15-puzzle domain. However, we have tried Micro-Hillary in several other domains. In all the domains, the initial states are generated using the domain-independent training-problem generator which applies a random sequence of operators to the goal state.
  - *24-puzzle:* $5 \times 5$ sliding tile puzzle.
  - *10-cannibals:* The cannibals and missionaries problem (Nilsson, 1980). To make it non-trivial, we use 10 cannibals and 10 missionaries instead of the usual 3 and 3. 10 cannibals and 10 missionaries are situated on the two banks of a river. They must all be moved onto one target bank. They can use a boat which can ship one or two persons. The cannibals should never outnumber the missionaries on either bank. The heuristic function used is the number of persons that are not yet located on the target bank.
  - *10-stones:* A puzzle that appeared in Nilsson's book (1980). Instead of using 3 black stones and 3 white stones we use 5 black stones and 5 white stones. The goal configuration is |W|W|W|W|W|B|B|B|B|B| |. A stone may move into an adjacent empty cell or may hop over one or two other stones into an empty cell. The heuristic function used is the number of stones not yet in their goal position.
  - *5-Hanoi:* The Towers of Hanoi problem with 5 rings in increasing sizes. There are 3 pegs and 5 rings. The goal position has the 5 rings placed on the left peg. A ring may be moved to the top of another peg provided that it is not placed on a smaller ring as a result. The heuristic function used is the number of rings not yet placed on their target peg.

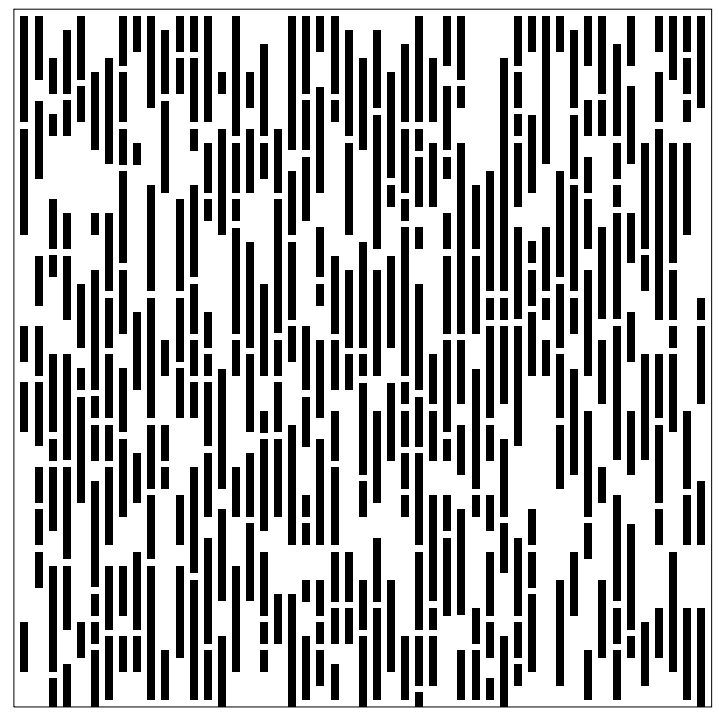

Figure 7: The grid domain

– *Grid:* A grid of $50 \times 50$ with random parallel walls inserted to make the Manhattan-distance heuristic inaccurate. There are four basic operators (North, South, West and East). A goal state can be any location on the grid. The objective is to find a path from the initial location to the goal location. Figure 7 shows the grid used.

The emphasis of this paper is on building a general learning algorithm that is able to solve the $N \times N$ puzzle problem. The goal of testing the algorithm on these other domains was to show that the learning algorithm is indeed general and does not include domain-specific procedures.

## 4.2 Learning to Solve Problems in the 15-puzzle Domain

| **Learning Resources** | Operator applications | | CPU seconds | | Problems | |
|---|---|---|---|---|---|---|
| | Mean | Std. | Mean | Std. | Mean | Std. |
| | 498,172 | 74,047 | 735 | 218 | 60 | 4.2 |
| **Generated Macros** | Total number | | Mean Length | | Max Length | |
| | Mean | Std. | Mean | Std. | Mean | Std. |
| | 14.16 | 0.74 | 8.38 | 0.2 | 18 | 0.0 |

Table 1: The upper part of the table shows the learning resources consumed while learning macros in the 15-puzzle domain. The results are averaged over 100 learning sessions. The lower part of the table lists the statistics of the generated macro sets (averaged over the 100 databases).

The basic experiment tests the ability of Micro-Hillary to learn the 15-puzzle domain. Table 1 shows the learning resources and the macro statistics averaged over the 100 learning sessions. We also tested the utility of the learned knowledge by comparing the performance of the problem solver that uses the learned macros to the following problem solvers:

1. Micro-Hillary without the learned knowledge and with learning turned off.
2. Best-first search[8] using the MD heuristic.
3. Best-first search using the RR heuristic.
4. Weighted-A* with the MD heuristic. We selected a node evaluation function $f(n) = (1-w)g(n) + wh(n)$ with $w = 0.75$, which was reported by Korf (1993) as the best weight for the 15-puzzle domain (Korf uses the equivalent notation $f(n) = g(n) + Wh(n)$ with W=3).
5. IDA* with the MD heuristic. Since we used random problems generated using the same method as Korf, we used the results reported in (Korf, 1993). The number of operator applications for Korf's data was estimated based on the ratio between generated nodes and operator applications in our run of best-first search.

Table 2 shows the results of this experiment. The results are averaged over the 100 problems in the test set. The results for Micro-Hillary after learning are also averaged over the 100 learning sessions.

| Search method | Operator applications | | Generated nodes | | Expanded nodes | | CPU seconds | | Solution length | |
|---|---|---|---|---|---|---|---|---|---|---|
| | Mean | Std. | Mean | Std. | Mean | Std. | Mean | Std. | Mean | Std. |
| Micro-Hillary (before learning) | 221,552 | 51,188 | 181,422 | 45,044 | 55,372.9 | 12,794.0 | 618.92 | 137.75 | 141.0 | 26.40 |
| Micro-Hillary (after learning) | 688 | 21 | 196 | 4 | 47.4 | 1.3 | 0.30 | 0.01 | 149.5 | 0.19 |
| Best first (MD) | 13,320 | 8,686 | 6,868 | 4,500 | 3,330.5 | 2,171.6 | 53.95 | 39.81 | 145.8 | 35.48 |
| Best first (RR) | >192,811 | 26,365 | >98,101 | 13,294 | >48,203 | 6591.4 | >5920 | 11,989 | | |
| WA* W=3 | 40,385 | 44,396 | 20,181 | 21,917 | 10,096 | 11099.0 | 439.28 | 1206.00 | 74.2 | 12.64 |
| IDA* Korf (1993) | 704,067,291 | | 363,028,090 | | | | | | 53.0 | |

Table 2: The utility of the learned macros. The resources required for solving the testing set compared with various non-learning problem solvers. The results are averaged over the 100 problems in the test set. The results for Micro-Hillary after learning are also averaged over the 100 learning sessions.

Micro-Hillary, after learning, solves problems much faster than the other algorithms. It uses 19 times fewer operator applications than best-first search using the MD heuristic, with solutions of comparable length. If we compare the CPU time of best-first search to

8. Best-first search has several meanings in the AI literature. Here we mean a heuristic search method that keeps the whole front of the search, expands the node with the best heuristic value and replaces it with its children. When two nodes have equivalent heuristic value, the one with the shorter distance from the initial state is prefered.

that of MICRO-HILLARY we can see that MICRO-HILLARY is more than 100 times faster than best-first search. This is because best-first search has an overhead per generated node which is much higher than MICRO-HILLARY's.[9] MICRO-HILLARY also has the advantage of a very small standard deviation. MICRO-HILLARY after learning is 322 times faster than MICRO-HILLARY before learning. MICRO-HILLARY is 59 times faster than WA* with $W = 3$, but yields solutions that are twice as long. MICRO-HILLARY is more than 1,000,000 times faster than IDA* with solutions that are 2.8 times longer (than the optimal solutions found by IDA*).

During the testing phase, MICRO-HILLARY never encountered local minima. Thus, according to definition 1, all the sets of macros learned by MICRO-HILLARY were apparently *complete*. This was also the case with the rest of the experiments described in this paper.

We have tried using iterative deepening (ID) for escape instead of ILB. We used an improved version of ID that tests each new state against the states along the path to the root to avoid loops. Table 3 shows the resources consumed during learning. Table 4 shows the statistics of the macro sets acquired. As expected, learning with ILB is faster than with ID (by a factor of 9), but produces macros that are slightly longer (ID finds the shortest escape routes).

| | Operator applications | | CPU seconds | | Problems | |
|---|---|---|---|---|---|---|
| Escape method | Mean | Std. | Mean | Std. | Mean | Std. |
| ILB | 498,172 | 74,047 | 735 | 218 | 60 | 4.2 |
| ID | 4,553,319 | 767,093 | 2,542 | 457 | 60 | 5.4 |

Table 3: The learning resources consumed while learning in the 15-puzzle domain using various escape methods. Each number represents the mean over 100 learning sessions.

| | Total number | | Mean Length | | Max Length | |
|---|---|---|---|---|---|---|
| Escape method | Mean | Std. | Mean | Std. | Mean | Std. |
| ILB | 14.16 | 0.74 | 8.38 | 0.20 | 18.0 | 0.0 |
| ID | 14.39 | 0.88 | 8.12 | 0.22 | 17.0 | 0.0 |

Table 4: The statistics of the macro sets acquired with various escape methods. Each number represents the mean over 100 databases.

### 4.3 The Effect of the Selection Method on Learning

We tested MICRO-HILLARY with the minimum-to-minimum and any-to-better selection strategies. The learning resources consumed are shown in Table 5. The statistics of the

9. We used a fast implementation of best-first search with efficient data structures: a heap and a hash table for the OPEN list, a hash table for the CLOSED list.

acquired macro sets are shown in Table 6. The performance of the problem solver using these macros is shown in Table 7.

| | Operator applications | | CPU seconds | | Problems | |
|---|---|---|---|---|---|---|
| Selection method | Mean | Std. | Mean | Std. | Mean | Std. |
| Minimum-to-better | 498,172 | 74,047 | 735 | 218 | 60 | 4.2 |
| Minimum-to-minimum | 22,211,450 | 27,096,382 | 5,461 | 5,298 | 426 | 434.0 |
| Any-to-better | 572,622 | 117,330 | 618 | 362 | 65 | 9.3 |

Table 5: The learning resources consumed while learning in the 15-puzzle domain using various selection methods. Each number represents the mean over 100 learning sessions.

| | Total number | | Mean Length | | Max Length | |
|---|---|---|---|---|---|---|
| Selection method | Mean | Std. | Mean | Std. | Mean | Std. |
| Minimum-to-better | 14.16 | 0.74 | 8.38 | 0.20 | 18.0 | 0.0 |
| Minimum-to-minimum | 104.04 | 98.92 | 109.90 | 90.38 | 348.4 | 172.8 |
| Any-to-better | 42.06 | 4.43 | 8.20 | 0.15 | 18.0 | 0.3 |

Table 6: The statistics of the macro sets acquired with various selection methods. Each number represents the mean over 100 databases.

| | Operator applications | | Expanded nodes | | CPU seconds | | Solution length | |
|---|---|---|---|---|---|---|---|---|
| Selection | Mean | Std. | Mean | Std. | Mean | Std. | Mean | Std. |
| Minimum-to-better | 688 | 21 | 47.4 | 1.3 | 0.30 | 0.01 | 149.5 | 0.19 |
| Minimum-to-minimum | 1281 | 332 | 43.5 | 2.1 | 0.84 | 0.12 | 251.5 | 50.08 |
| Any-to-better | 1396 | 132 | 46.4 | 0.7 | 0.39 | 0.03 | 143.8 | 3.20 |

Table 7: The utility of the learned macros with various selection methods. The results are first averaged over the 100 problems in the test set and then averaged again over the 100 macro sets.

These results confirm our analysis in the previous section. The minimum-to-minimum strategy acquires macros that are unnecessarily long, causing a large increase in the learning resources required. Also, because the macros are longer, they are less general and many of them are acquired before quiescence. The any-to-better strategy acquires unnecessary macros, which leads to a larger branching factor and lower utility.

### 4.4 The Effect of the Heuristic Function on Learning

The Micro-Hillary algorithm is based upon the availability of a "generally good" heuristic function to start with. The heuristic function should be able to lead the search program towards the goal, except in a small number of local minima where an exhaustive search and learning take place. The next experiment tests the effect of the heuristic function on the performance of Micro-Hillary. We have conducted 100 learning sessions and testing sessions with each of the heuristic functions described in Section 4.1.2. Tables 8, 9 and 10 show the results obtained.

| | Operator applications | | CPU seconds | | Problems | |
|---|---|---|---|---|---|---|
| Heuristic | Mean | Std. | Mean | Std. | Mean | Std. |
| RR | 498,172 | 74,047 | 735 | 218 | 60 | 4.2 |
| Row-by-row-2 | 4,391,468 | 1,805,776 | 2191 | 1,181 | 247 | 62.9 |
| Reduction | 552,463 | 128,769 | 878 | 334 | 63 | 9.1 |
| Spiral | 891,728 | 144,633 | 1,621 | 47 | 74 | 15.3 |
| MD | 228,487,090 | 44,106,048 | 98,659 | 2,341 | 1,917 | 210.3 |

Table 8: The learning resources consumed while learning in the 15-puzzle domain using various heuristic functions. Each number represents the mean over 100 learning sessions.

| | Total number | | Mean Length | | Max Length | |
|---|---|---|---|---|---|---|
| Heuristic | Mean | Std. | Mean | Std. | Mean | Std. |
| RR | 14.16 | 0.74 | 8.38 | 0.20 | 18.0 | 0.0 |
| Row-by-row-2 | 73.44 | 3.65 | 5.60 | 0.14 | 27.7 | 2.1 |
| Reduction | 16.59 | 0.76 | 6.08 | 0.08 | 13.0 | 0.0 |
| Spiral | 24.10 | 1.04 | 6.43 | 0.09 | 13.0 | 0.0 |
| MD | 730.00 | 40.09 | 16.37 | 0.23 | 48.2 | 1.0 |

Table 9: The statistics of the macro sets acquired with various heuristic functions. Each number represents the mean over 100 databases.

The row-by-row-2 function is a crippled version of the RR heuristic. It is therefore not surprising that the performance degraded by all measures when using it. It is more surprising to note that the performance also degraded with the powerful MD heuristic, which, according to Table 2, is much better than RR even for satisficing search. The lower performance is expressed in two main differences:

1. The learning resources with the MD heuristic are 460 times higher than with the RR heuristic.

2. The utility of the learned macro sets is lower—the number of operator applications is 16 times higher with the MD heuristic.

| | Operator applications | | Expanded nodes | | CPU seconds | | Solution length | |
|---|---|---|---|---|---|---|---|---|
| Heuristic | Mean | Std. | Mean | Std. | Mean | Std. | Mean | Std. |
| RR | 688 | 21 | 47.4 | 1.3 | 0.30 | 0.01 | 149.5 | 0.19 |
| Row-by-row-2 | 2,827 | 21 | 57.24 | 144.8 | 1.33 | 3.60 | 148.2 | 0.19 |
| Reduction | 738 | 19 | 42.67 | 0.4 | 0.29 | 0.01 | 125.5 | 1.60 |
| Spiral | 944 | 34 | 44.76 | 0.5 | 0.37 | 0.02 | 130.1 | 5.27 |
| MD | 11,520 | 1009 | 175.00 | 83.6 | 2.40 | 0.01 | 142.0 | 2.04 |

Table 10: The utility of the learned macros with various heuristic functions. The resources required for solving the testing set compared with various non-learning problem solvers. The results are first averaged over the 100 problems in the test set and then averaged again over the 100 macro sets.

To understand the reasons for these differences we can look at the statistics of the acquired macros (Table 9). The total number of macros acquired with MD is 50 times larger than the number of macros acquired with RR. A larger number of macros implies a larger branching factor, which explains the lower efficiency of the problem solver using the MD macros. The high learning resources can be explained by three factors:

1. The problem solver is used during training. Therefore, lower efficiency of the problem solver implies slower learning.

2. The larger number of macros indicates a larger number of training problems before quiescence. Indeed, with the MD heuristic, MICRO-HILLARY solved 30 times more problems during training.

3. The mean length of the macros acquired with the MD heuristic is twice as large as the mean length of macros acquired with the RR heuristic. Even more significant is the difference in the maximal length of a macro. With the MD heuristic MICRO-HILLARY acquires macros that are 48 operators long, while the maximal length of a macro using the RR heuristic is only 18. This indicates that the maximal radius of the MD heuristic is much larger than that of the RR heuristic. Larger radius implies more resources invested in escaping from local minima.

The above explanation is only partially satisfactory, since it does not tell us the *source* of the problem. To understand why the MD heuristic has a large radius, look at the following puzzle:

| | | | | |
|---|---|---|---|---|
| 2 | 1 | 3 | 4 | 5 |
| 6 | 7 | 8 | 9 | 10 |
| 11 | 12 | 13 | 14 | 15 |
| 16 | 17 | 18 | 19 | 20 |
| 21 | 22 | 24 | 23 | |

The sum-of-Manhattan distances for this puzzle is only 4, yet a very long sequence of moves is required to reduce the distance. The larger radius implies longer (and less general) macros and more resources for the escape processes.

| MD | | | | | | | | | RR | | | | | | | | |
|---|---|---|---|---|---|---|---|---|---|---|---|---|---|---|---|---|---|
| Start | | | | Macro | End | | | | Start | | | | Macro | End | | | |
| 2 | 1 | 4 | 8 | r r u u u | 2 | 1 | 4 | | 1 | 2 | 3 | 4 | d l l u r | 1 | 2 | 3 | 4 |
| 13 | 6 | 15 | 12 | | 13 | 6 | 15 | 8 | 5 | 6 | 7 | 8 | | 5 | 6 | 7 | 8 |
| 5 | 11 | 3 | 10 | | 5 | 11 | 3 | 12 | 9 | 13 | 10 | | | 9 | 10 | | 12 |
| 9 | | 7 | 14 | | 9 | 7 | 14 | 10 | 11 | 14 | 15 | 12 | | 11 | 13 | 14 | 15 |
| | 2 | 3 | 4 | r d l d r u u l d r d | 1 | 2 | 3 | 4 | 1 | 2 | 3 | 12 | l u r r d l u l d | 1 | 2 | 3 | 4 |
| 5 | 9 | 7 | 8 | | 9 | 5 | 7 | 8 | 8 | 15 | | 4 | | 8 | | 12 | 15 |
| 1 | 10 | 11 | 12 | | 10 | | 11 | 12 | 14 | 5 | 13 | 6 | | 14 | 5 | 13 | 6 |
| 13 | 6 | 15 | 14 | | 13 | 6 | 15 | 14 | 9 | 11 | 7 | 10 | | 9 | 11 | 7 | 10 |
| 1 | 2 | 3 | 4 | u l u r d l l l u r r r d d l l u r u l l r d d r r l r u d l l l u r u l d d | 1 | 2 | 3 | 4 | 1 | 2 | 3 | 4 | u l d l l u r d r u l l d r r u r d | 1 | 2 | 3 | 4 |
| 9 | 6 | 8 | 7 | | 5 | 6 | 7 | 8 | 5 | 6 | 7 | 8 | | 5 | 6 | 7 | 8 |
| 5 | 10 | 11 | 12 | | 9 | 10 | 11 | 12 | 9 | 10 | 11 | 12 | | 9 | 10 | 11 | 12 |
| 13 | 14 | 15 | | | | 13 | 14 | 15 | 14 | 15 | 13 | | | 15 | 13 | 14 | |

Table 11: Examples for macros acquired with the RR and the MD heuristics. Each example shows the state where the macro was acquired, the macro itself, and the state after the application of the acquired macro.

It is more difficult to explain the larger number of macros. A larger number of macros means that during training Micro-Hillary encounters many states where none of the previously learned macros is useful, and new macros are acquired. This implies that the likelihood of a macro acquired with MD to be useful in a state is much lower than that of a macro acquired with RR. One reason for this difference is the longer macros acquired with MD. As we stated before, longer macros tend to be less general. There are, however, other reasons which are best understood by examples. Table 11 shows three macros acquired with each of the two heuristics, along with the states before and after the macro acquisition.

The MD macro at the first line, *rruuu*, is applicable only in states where the empty tile is in the last row at one of the two leftmost locations. Five tiles are moved and the Manhattan distance of each is either increased or decreased by one. The macro is useful only in states where the distance of at least three of the tiles is decreased. Compare this to the equally long RR macro *dllur*. It is applicable in states where the empty tile is in one of 6 locations (last two columns and last three rows), and is therefore much more likely to be applicable than the MD macro. It is useful in states where the blank is to the right of the next tile to be moved and the direction of the movement is to the left. Since the RR heuristic imposes a particular order on the placement of tiles, *any* attempt to solve a problem will get to a state where, for example, the first 9 tiles are in place and tile 10 should be placed next. At a certain point tile 10 will be at distance 1 from its target location, either below the target location or to the right of it. In all the states where the tile is to the right of its target location, and the blank tile is to its right, the macro *dllur* will be useful. Thus, this macro has high likelihood of being useful.

The MD macro in the second line is, again, useful for a very particular situation. The RR macro, however, is useful for at least half of the problems. While solving a problem with RR there must be a situation where the first 3 tiles are in their target location and the fourth tile is at distance one from its target. If the blank is to its left, then the macro is useful. The MD macro in the last row of the table is the most extreme case of low usefulness. It is useful only if the tiles are placed in a particular order, with the tiles in the first, third and fourth row placed before the tiles in the second row. Contrast this macro with the longest RR macro shown to its right. This macro is useful for most problems since there will always be a situation where the first three rows are ordered and tile 13 should be moved towards its goal.

To conclude, the difference between RR and MD that makes the RR macros more useful is the restriction that RR puts on the order of solving subproblems. This strict order makes the problem solver go through similar states regardless of the initial state. The freedom given by MD allows the problem solver pick any of the $N^2!$ possible orders for tile placement. Furthermore, it allows macros that move tiles away from their target location, provided that other tiles are moved towards their target. Such macros are useful only in very particular states.

The *particular* order used by RR reduces the maximal radius by making it possible to place tiles without the need to (even temporarily) displace too many already-placed tiles. The other two heuristics, Reduction and Spiral, also have this property. Consider, however, a heuristic similar to RR where the tiles are placed in order from tile 2 to tile 15 after which tile 1 is placed. Such a placement order increases the maximal radius significantly. In Section 6 we discuss the differences between the two heuristics when used for the $N \times N$ puzzle domain.

### 4.5 Learning Macros in Other Domains

| | Operator applications | | CPU seconds | | Problems | |
|---|---|---|---|---|---|---|
| Domain | Mean | Std. | Mean | Std. | Mean | Std. |
| 24-puzzle | 859,497 | 186,823 | 4842.0 | 2270.0 | 62.8 | 7.2 |
| 10-Cannibals | 216,572 | 88,467 | 14.2 | 6.0 | 63.9 | 12.0 |
| 10-Stones | 144,303 | 1,653 | 8.4 | 1.9 | 52.0 | 0.3 |
| 5-Hanoi | 377,671 | 80,968 | 35.2 | 3.5 | 76.2 | 9.5 |
| Grid | 1,956,972 | 1,191,383 | 58.8 | 35.0 | 185.0 | 60.2 |

Table 12: The resources consumed while learning in various domains. Each number represents the mean over 100 learning sessions.

We have applied MICRO-HILLARY to the other domains specified in Section 4.1. Tables 12, 13 and 14 show the mean results for 100 learning sessions. MICRO-HILLARY was able to reach quiescence in all the domains. The 10-stones and 10-cannibals domains are very simple. One or two macros were sufficient to reach quiescence. Note that we used the same quiescence parameter, 50 problems, for all the domains. After solving each problem, Micro-Hillary increases by 100 the length of the random sequence used for generating a training problem. Therefore, Micro-Hillary spends 125,000 operator applications just to make sure that there is nothing new to learn. In the simple domains, this amounts to most of the resources used by Micro-Hillary.

| | Total number | | Mean Length | | Max Length | |
|---|---|---|---|---|---|---|
| Domain | Mean | Std. | Mean | Std. | Mean | Std. |
| 24-puzzle | 15.32 | 0.79 | 8.68 | 0.24 | 18.00 | 0.0 |
| 10-cannibals | 2.16 | 0.39 | 2.93 | 0.15 | 3.98 | 0.2 |
| 10-stones | 1.20 | 0.40 | 2.00 | 0.00 | 2.00 | 0.0 |
| 5-Hanoi | 11.47 | 0.71 | 7.24 | 0.25 | 16.00 | 0.0 |
| Grid | 17.50 | 3.14 | 8.62 | 0.82 | 16.36 | 1.5 |

Table 13: Statistics of the macro sets generated in various domains. Each number represents the mean over 100 sets.

| | Before learning | | After learning | |
|---|---|---|---|---|
| Domain | Mean | Std. | Mean | Std. |
| 24-puzzle | 711,545 | 134,807.0 | 1540.0 | 57.5 |
| 10-cannibals | 57 | 40.5 | 31.6 | 0.4 |
| 10-stones | 206 | 84.8 | 126.6 | 8.1 |
| 5-Hanoi | 10,993 | 14,623.3 | 156.0 | 9.4 |
| Grid | 1,529 | 1313 | 369.0 | 35.0 |

Table 14: The performance of Micro-Hillary (in operator applications) before and after learning in various domains.

It is interesting to look at the macros learned in the grid domain. Most of the macros have a structure of *sss...swn...nnn*, where $S$ stands for south, $W$ for west, $N$ for north, and $S$ and $N$ are equal in number. Such macros are used to make detours around walls that block the search.

Micro-Hillary was able to improve the performance of problem solving in each of the domains. The most notable improvement is in the 24-puzzle domain where the performance after learning is 462 times better than the performance before learning.

## 5. Solving the General $N \times N$ Puzzle

Some domains can be scaled up or down by adjusting a certain parameter, such that the same operators are applicable in the scaled domain (perhaps by using the parameter in their definition). For example, we can define a family of domains called the $N \times N$ puzzle, where each $N$ defines a different puzzle domain. When $N = 3$ we get the 8-puzzle domain, when $N = 4$ we get the 15-puzzle domain, etc. All the domains in this family use the same operators: *Up*, *Down*, *Left* and *Right*.

### 5.1 Using Micro-Hillary in Scalable Domains

In the previous section we showed how Micro-Hillary can be used to efficiently solve problems in the 15-puzzle and the 24-puzzle domains. Is it possible to use Micro-Hillary to efficiently solve general $N \times N$ sliding-tile puzzle problems? We have developed a learning algorithm called Parametric Micro-Hillary that uses Micro-Hillary in scalable domains. The original Micro-Hillary gets a set of basic operators and a well-behaved heuristic function. Parametric Micro-Hillary gets in addition the name of the domain parameter and an initial value for it. The basic operators and the heuristic function can use the domain parameter in their definition. The algorithm sets the parameter to its initial value and calls Micro-Hillary. When Micro-Hillary returns (due to quiescence), the parameter is increased and Micro-Hillary is called again. The algorithm stops when no new macros are added. The algorithm is shown in Figure 8.

```
procedure Parametric Micro-Hillary(BasicOps,h,InitialParameterValue)
    Macros ← {}
    DomainParameter ← InitialParameterValue
    Loop
        MacrosBefore ← Macros
        Macros ← Micro-Hillary(BasicOps,h)
; BasicOps and h can use DomainParemeter in their definition
        DomainParameter ← DomainParameter + 1
    Until MacrosBefore = Macros
    Return Macros
```

Figure 8: The Parametric Micro-Hillary algorithm

The idea behind this algorithm is to extend the strategy of generating training problems of increasing difficulty. Micro-Hillary increases the length of the random sequences to generate more complex problems, while Parametric Micro-Hillary also increases the domain parameter.

### 5.2 Experimenting with Parametric Micro-Hillary in the $N \times N$ Puzzle Domains

We have experimented with the Parametric Micro-Hillary algorithm in the $N \times N$ puzzle domains using the above specifications. The resulting macro sets were tested on random $10 \times 10$ puzzles.

Table 15 shows the summary of 100 learning sessions. Micro-Hillary learned most of its macros by solving $3 \times 3$ and $4 \times 4$ puzzles. It typically learned one or two macros while solving $5 \times 5$ puzzles, and reached quiescence while solving $6 \times 6$ problems without learning any new macros. Note that on average at least 85% of the training problems and at least half of the learning time (operator applications) were used just for testing quiescence. Although Parametric Micro-Hillary solved 232 training problems on average compared with an average of 60 problems solved by Micro-Hillary in the 15-puzzle domain, the total learning CPU time is almost 3 times shorter than the time spent for learning the 15-puzzle. This is an indication that the method of solving problems in increasing level of difficulty indeed saves learning resources. Parametric Micro-Hillary was able to learn part of the macros while solving $3 \times 3$ puzzles, thus saving time by escaping local minima in a domain with a lower complexity.

| Learning Resources | Operator applications | | CPU seconds | | Problems | |
|---|---|---|---|---|---|---|
| | Mean | Std. | Mean | Std. | Mean | Std. |
| | 1084571 | 126338 | 270.7 | 62.7 | 232.8 | 13 |
| **Generated Macros** | Total number | | Mean Length | | Max Length | |
| | Mean | Std. | Mean | Std. | Mean | Std. |
| | 14.87 | 0.83 | 8.82 | 0.23 | 18 | 0 |
| **Performance** | Operator applications | | CPU seconds | | Solution Length | |
| | Mean | Std. | Mean | Std. | Mean | Std. |
| | 15891 | 1020 | 15.64 | 0.85 | 3028 | 268 |

Table 15: The upper part of the table lists the learning resources consumed by learning macros in the $N \times N$ puzzle domain. The middle part of the table contains the statistics of the generated macro sets. The lower part shows the performance of Micro-Hillary while using the macros to solve 100 random $10 \times 10$ puzzle problems. All the reported data is averaged over 100 learning sessions.

A second test was performed with one of the macro sets (we arbitrarily selected the one that was learned first). We generated sets of 10 random testing problems for various sizes up to $50 \times 50$ and solved these problems using the macro set. The random problems were generated, as in the previous section, by even random permutations. Figure 9 shows the mean number of operator applications as a function of the puzzle size. Micro-Hillary solved the whole test set in a reasonable time. It looks as if the program can now efficiently solve any $N \times N$ solvable problem. We have indeed succeeded in proving that the set of macros learned is *complete* and that Micro-Hillary can solve any solvable problem in $O(N^3)$ with a reasonable constant (288). The proof is given in the appendix. This proof is supported by our empirical testing. Micro-Hillary never resorted to the escape procedure throughout the whole testing phase. The proof is for a specific macro set, but can be repeated for the other macro sets as well. A variation of that proof shows that Micro-Hillary can solve any solvable problem in $O(N^3)$ even without learning. However, the constant in this upper limit is prohibitively high ($10^{12}$). We do not claim that Micro-Hillary will always perform successful learning. It is possible that Micro-Hillary will reach quiescence and quit learning, having missed essential macros. The probability of such an event, however, diminishes with an increase in the size of the quiescence parameter.

Note that both the empirical results and the formal proof show that the maximal radius of any puzzle problem, regardless of the puzzle size, with respect to the rr heuristic, is bounded by 18. This bounded-radius property is a necessary condition for Micro-Hillary to scale up its acquired macros for domains with a larger parameter. It can be easily shown that the MD heuristic does not have such a property.

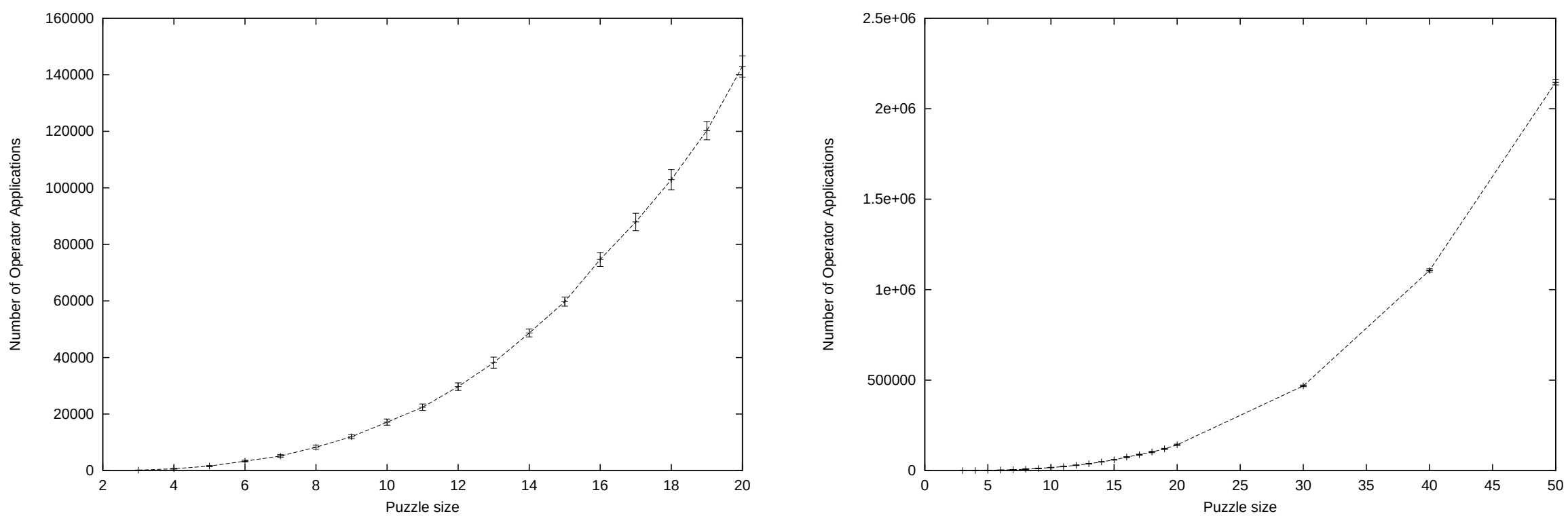

(a) (b)

Figure 9: Testing the macros learned by Parametric Micro-Hillary on large puzzles. The graphs show mean number of operator applications as a function of the puzzle size. The error bars are one standard deviation away on either side of the mean. (a) Results for sizes up to 20. (b) Results up to size 50.

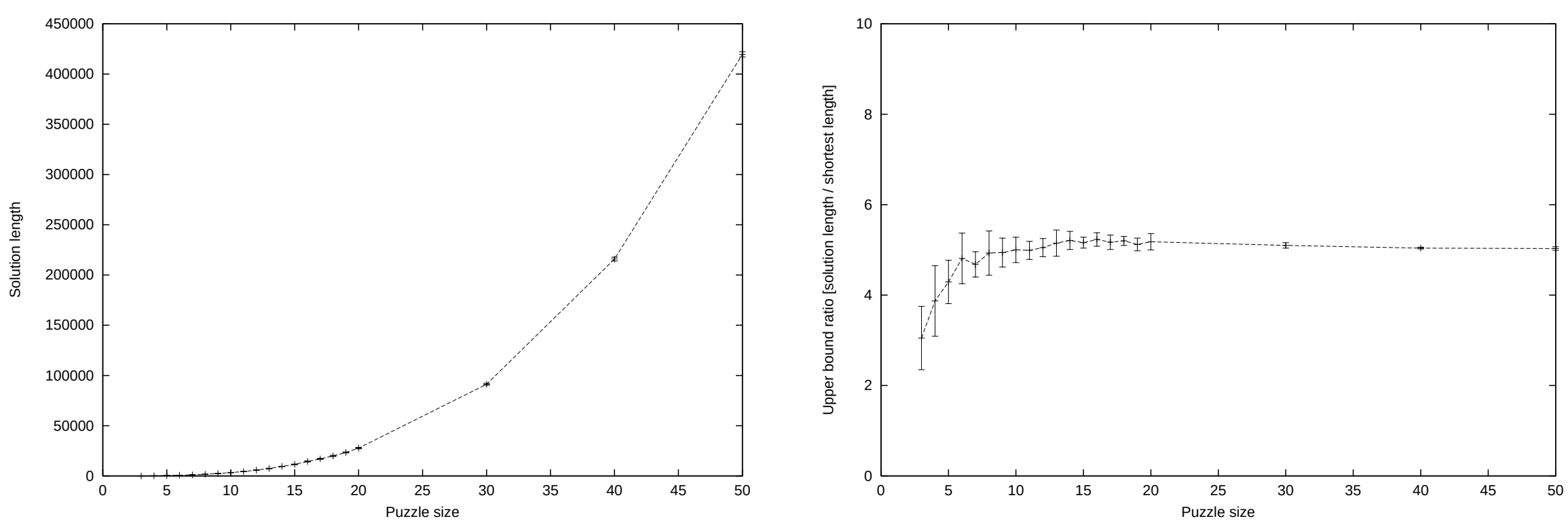

(a) (b)

Figure 10: (a) The solution length (in basic moves) as a function of the puzzle size. (b) The ratio between the solution length and the heuristic lower-bound on the shortest solution as a function of the puzzle size. In both graphs the error bars are one standard deviation away on either side of the mean.

Figure 10(a) shows the mean solution length as a function of the puzzle size. Figure 10(b) shows the upper bound on the ratio of the solution length to the optimal solution length as a function of the puzzle size (the solution length divided by the sum-of-Manhattan-distances). It is interesting to note that the graph is flattened at a value of 5. Ratner and Warmuth (1986, 1990) sketch a (quite complicated) hand-crafted approximation algorithm for the $N \times N$ puzzle. In their earlier paper they give a constant factor of 22 to the approximation. Looking at Figure 10, it is quite possible that Micro-Hillary could have found such an algorithm by itself. The constant factor of Micro-Hillary seems to be close to 5, but it is measured with respect to *random* problems, while Ratner and Warmuth's constant is a worst-case upper bound. In the appendix, we prove that the length of the solutions found by Micro-Hillary for $N \times N$ puzzle problems is $O(N^3)$ with a constant of 50 (which is substantially worse than Ratner and Warmuth's).

### 5.3 Experimenting with Parametric Micro-Hillary in Other Scalable Domains

We tested Parametric Micro-Hillary in other parameterized domains. In the N-Cannibals and N-Stones domains, Parametric Micro-Hillary learned all the macros with the minimal value of the parameter (3). The test was performed using problems with a parameter of 20 in both domains. Micro-Hillary's performance indeed improved in both domains and problem solving proceeded without encountering local minima.

The N-Hanoi domain family is recursive in nature and we did not expect Micro-Hillary to find a complete macro set for these domains. The length of the macros should grow with the number of rings; therefore, Micro-Hillary should not reach quiescence in these domains. We were surprised to find that Parametric Micro-Hillary achieved quiescence after solving problems of 7 or 8 rings (7.13 on average). This error was caused by the domain-independent training-problems generator. The probability that the largest ring will be moved from its target location after a random sequence of moves is very low. Indeed, when we increased the length of the sequences used for generating training problems and increased the quiescence parameter, learning continued, but Micro-Hillary still reached quiescence after solving problems with 9 rings. In both cases, the macros learned were not sufficient for solving problems with a parameter that is larger than the values encountered during training. The test was performed with problems of 6 rings. For solving domain families such as N-Hanoi, we should extend Micro-Hillary and endow it with the capability of generating recursive macros. To avoid the problem of Parametric Micro-Hillary quitting prematurely, we can modify the problem generator of Micro-Hillary to use random sequences with lengths that are based on the domain parameter. Alternatively, we can use domain-specific problem generators. The results of this experiment are summarized in Tables 16, 17 and 18.

| | Operator applications | | CPU seconds | | Problems | |
|---|---|---|---|---|---|---|
| Domain | Mean | Std. | Mean | Std. | Mean | Std. |
| N-Cannibals | 331,183 | 59723 | 20.84 | 3.8 | 112 | 8.5 |
| N-Stones | 277,852 | 3384 | 11.53 | 2.7 | 1030 | 0.6 |
| N-Hanoi | 2,427,186 | 508,093 | 579.00 | 176.0 | 370 | 43.7 |

Table 16: Summary of the learning resources consumed in various domains. Each number represents the mean over 100 learning sessions.

| Domain | Total number | | Mean Length | | Max Length | |
|---|---|---|---|---|---|---|
| | Mean | Std. | Mean | Std. | Mean | Std. |
| N-Cannibals | 4.40 | 0.5 | 2.4 | 0.05 | 4.0 | 0.0 |
| N-Stones | 1.22 | 0.4 | 2.0 | 0.00 | 2.0 | 0.0 |
| N-Hanoi | 16.08 | 0.9 | 12.5 | 1.24 | 33.2 | 6.6 |

Table 17: Statistics of the macro sets generated in various domains. Each number represents the mean over 100 sets.

| Domain | Before learning | | After learning | |
|---|---|---|---|---|
| | Mean | Std. | Mean | Std. |
| N-Cannibals | 150 | 85 | 105 | 10 |
| N-Stones | 3,671 | 605 | 2,009 | 155 |
| N-Hanoi | 171,956 | 168,338 | 2,913 | 16,530 |

Table 18: The performance of MICRO-HILLARY before and after learning in various domains.

Another related experiment involved transfer of knowledge between two similar domains (but not parameterized as the domains above). We generated a random grid, different from the one used for the experiments above, and performed a testing session with MICRO-HILLARY, using macros that were learned in the first grid. Using the macros improved MICRO-HILLARY's performance—from 1529 operator applications without macros down to 594 with macros. This ability of transferring skill from one grid to another arises from the similar shape of obstacles. A macro such as SSSSSWNNNNN can be helpful in getting around walls of various sizes in the original grid used for learning and in the grid used for testing.

## 6. Discussion

Despite its simplicity, the MICRO-HILLARY algorithm presented in this paper was able to learn macros that allow an efficient solution of any solvable $N \times N$ puzzle problem. It also performed well in a number of other domains. In this section, we compare MICRO-HILLARY to other macro-learning algorithms and discuss its strengths and weaknesses.

Most of the existing macro-learning programs are based on the notion of subgoaling: the learner tries to acquire macros that achieve some subgoal without undoing previously satisfied subgoals (Korf, 1985; Laird et al., 1986; Ruby & Kibler, 1989, 1992; Tadepalli, 1991; Tadepalli & Natarajan, 1996). MICRO-HILLARY, like MACLEARN (Iba, 1985), does not assume subgoaling, but assumes the existence of a heuristic function. EASe (Ruby & Kibler, 1992) combines subgoaling with a heuristic function to guide the search for the current subgoal. The subgoal-oriented macro-learners use various methods to guard the previously achieved subgoals. MICRO-HILLARY is much simpler, requiring only that the macro acquired lead from a local minimum to a state with a better heuristic value. It is possible to implicitly enforce subgoaling by using an appropriate evaluation function. For example, the RR heuristic used for the $N \times N$ puzzle domain encodes a subgoal-oriented strategy for solving puzzle problems. However, the algorithm works even in domains where such natural subgoaling is not evident. For example, in the grid domain, the local minima are located behind obstacles—the search procedure cannot move forward and all other directions increase the Manhattan distance from the goal location. Micro-Hillary acquires macros that allow a detour around the obstacles. Such local minima cannot be called subgoals since we would prefer to avoid them. However, once the search program is trapped in those minima, it can use the macros to escape from them.

While Micro-Hillary and MacLearn are similar in their use of the heuristic function as a guide to the macro learner, they are different in several other aspects. MacLearn uses best-first search as its training and testing problem solver. Micro-Hillary uses hill-climbing (with an escape procedure). This difference is significant since the increased branching factor resulting from the macros is extremely harmful in best-first search but adds only to the constant in hill-climbing search (provided that the hill-climbing search does not find solutions that are significantly longer than those found by best-first search). Related to this difference is the different approaches that MacLearn and Micro-Hillary take toward generalization of macros. MacLearn, using best-first search, cannot afford trying every macro in every state. Therefore, it uses domain-specific pattern language to specify for each macro the states at which it should be applied. Micro-Hillary, using hill-climbing, can afford more liberal and domain-independent generalization using a macro wherever it is legal. Another difference between the two algorithms is the attention filters that they use. MacLearn uses the minimum-to-minimum method while Micro-Hillary uses the minimum-to-better method. The minimum-to-minimum method has the weakness of acquiring longer macros, as local minima become scarce and far apart during the learning process. MacLearn needed a special filter for avoiding the acquisition of long macros.

Micro-Hillary is a completely autonomous algorithm—it generates its own examples and uses the concept of *quiescence* to decide when to increase the domain parameter and when to stop learning. MacLearn used a more supervised approach, processing hand-crafted training examples. To learn to solve sliding-tile puzzles, MacLearn was given first a single problem of a 6-tile puzzle, then a single instance of an 8-puzzle and a 15-puzzle (after which it was able to solve a single instance of a 24-puzzle). Micro-Hillary was demonstrated solving large sets of random problems of large size. One advantage that MacLearn has over Micro-Hillary is its ability to handle domains with parameterized operators. Micro-Hillary tries all the applicable macros in every state. Had parameterized macros been used, Micro-Hillary would have had to try all the ways of binding the parameters, resulting in a potentially large branching factor. The peg-solitaire domain cannot be handled efficiently by Micro-Hillary since it does not use a small set of fixed operators. MacLearn solves it by using parameterized operators.

Unlike some speedup learners that provide us with either statistical or theoretical guarantees (Cohen, 1992; Gratch & DeJong, 1992; Greiner & Likuski, 1989; Subramanian & Hunter, 1992; Tadepalli & Natarajan, 1996), Micro-Hillary has a heuristic nature and does not provide us with any guarantee. Indeed, while it performs very well in some domains, it fails in other domains such as the N-Hanoi. To handle such domains, we would have to endow Micro-Hillary with the capability of learning parameterized recursive macros (Shavlik, 1990). A related weakness of MICRO-HILLARY is the sensitivity of the algorithm to the heuristic function available. We have shown an example where MICRO-HILLARY has extreme difficulties with a seemingly good heuristic (the sum of Manhattan distances for the 15-puzzle) while learning easily with another one.

There are two features of the heuristic function that are necessary for a successful performance of MICRO-HILLARY:

- There is a "small" set of operator sequences that completely smooth the heuristic function to eliminate any local minima. This can be achieved either by a heuristic function that has a small number of local minima or, when the set of local minima is large, by macros that generalize well over this set.
- The maximal *radius* of the local minima should be small enough to allow the escape procedure to find an escape route. In addition, to use the scalable version of MICRO-HILLARY, the maximal radius should not increase with the size.

The experiments described in this paper and the theorem in the appendix show that the RR heuristic of the $N \times N$ puzzle domain has a maximal radius of 18, regardless of the puzzle size. This feature does not hold for the MD heuristic. The example puzzle in Section 4.4 illustrates the problem. When scaling up the puzzle, the maximal radius scales up as well.

Another weakness of MICRO-HILLARY is its sensitivity to the quiescence parameter. This parameter tells MICRO-HILLARY when to quit learning. For PARAMETRIC MICRO-HILLARY it is used to determine when to increase the domain parameter for training. Setting the quiescence parameter to a low value can make MICRO-HILLARY quit prematurely before learning all the necessary macros. Setting it to a high value can lead to a great waste of learning resources. Indeed, for the simple domains such as the N-Cannibals and N-Stones, MICRO-HILLARY spent most of its learning time making sure that there is nothing new to learn. Note that the quiescence mechanism is needed only when performing *off-line* learning. MICRO-HILLARY could also be used in an on-line mode, where macros are learned while solving user problems. In such a case there is no need to use the quiescence parameter. Whenever MICRO-HILLARY encounters a local minimum, it will use the escape mechanism to find a path and record it as a macro.

While the above weaknesses are significant, we should also consider the strengths of the algorithm. MICRO-HILLARY is extremely simple, yet it is able to learn to efficiently solve problems in a number of known domains. The ability of such a simple learning algorithm to find an efficient procedure for solving general $N \times N$ puzzle problems shows the potential in using selective learning for speeding up problem solvers. Our research goal of building the simplest domain-independent learning algorithm to solve the $N \times N$ puzzle was achieved. A Common-Lisp implementation of MICRO-HILLARY and the $N \times N$ domain is included in the online appendix. This short code (less than 200 lines long) underscores the simplicity of the algorithm.

Even with the simple architecture of MICRO-HILLARY, there is much future work to be done. One of the interesting directions is the application of macro-learning to more complex domains such as scheduling. Macro-learning, like most learning techniques, works best when there are regularities in the domain that it can exploit. It is not yet clear whether there are such regularities in the scheduling domain, for example. It is possible that macros can encode groups of variable assignments that should be tried together.

| | | | | |
|---|---|---|---|---|
| 1 | 2 | 3 | 4 | 5 |
| 6 | 7 | 8 | 19 | 16 |
| 14 | | 17 | 18 | 15 |
| 20 | 24 | 13 | 22 | 21 |
| 23 | 9 | 10 | 11 | 12 |

$s_g = \langle 1, 2, 3, 4, \ldots, 23, 24, 0 \rangle$

| | | |
|---|---|---|
| $tile_s(0,0) = 1$ | $tile_s(0,4) = 5$ | $tile_s(4,4) = 12$ |
| $loc_s(4) = (0,3)$ | $loc_s(0) = (2,1)$ | $loc_s(12) = (4,4)$ |
| $placed(s, s_g) = 8$ | $NextTile(s) = 9$ | $NextLoc(s) = (1,3)$ |

Figure 11: An example for the notations used for the $N \times N$ puzzle.

## Appendix A. Applying Parametric Micro-Hillary to the $N \times N$ Puzzle

To show a more concrete example of defining the parameters for Parametric Micro-Hillary we list here the exact definitions used to apply the algorithm to the $N \times N$ puzzle domains. The domain parameter used is $N$ with an initial value of 3.

To make the specifications more concise, we use the following definitions and notations. A *puzzle state* is a permutation of the sequence $\langle 0, 1, \ldots, N^2 - 1 \rangle$. Each of the elements of the puzzle state is called a *tile*. 0 is called the *empty tile*. For convenience, we will represent a puzzle state by an $N \times N$ row-major matrix. Let $s$ be a state, let $i, j \leq N$. We define $tile_s(i,j)$ to be the tile located in row $i$ column $j$ of $s$, where $s$ is represented by a row-major $N \times N$ matrix. For every state $s$ and tile $t \in s$, we define the tile location $loc_s(t) = (i,j)$ where $tile_s(i,j) = t$. Let $l_1 = (i_1, j_1)$ and $l_2 = (i_2, j_2)$ be two locations. The *distance* between $l_1$ and $l_2$ is defined as $d(l_1, l_2) = |i_1 - i_2| + |j_1 - j_2|$. Let $s = \langle t_1, \ldots, t_{N^2} \rangle$ be a state. Let $s_g = \langle g_1, \ldots, g_{N^2-1}, 0 \rangle$ be the goal state. The number of *placed* tiles is the largest $p$ such that $t_i = g_i$ for all $i \leq p$. The expression of $p + 1$ in the matrix notation is called *the next location* and is marked as $NextLoc(s)$. $g_p$ is called the *next tile* and is marked as $NextTile(s)$. Figure 11 shows examples for the above notations. The heuristic function is the one described in the beginning of Section 4, generalized to the $N \times N$ puzzle. It is a linear combination of three factors. The weights were chosen to be high enough to enforce a lexicographic order. The least significant factor is the Manhattan distance between the empty tile and the next tile. Since it is bounded above by $2(N-1)$, the second factor is multiplied by $2N$ to make sure that it is dominant. For the same reason, we multiply the most significant factor by $4N^2$.

Table 19 lists the exact parameters used for Micro-Hillary in the $N \times N$ puzzle domain.

## Appendix B. The Efficiency of Solving $N \times N$ Puzzles by Micro-Hillary

This section contains a proof that a specific set of macros learned by Micro-Hillary is *complete* with respect to the heuristic function $h$ defined in Table 19, and that a problem solver that uses these macros can solve any solvable $N \times N$ puzzle problem in $O(N^3)$ with a reasonable constant. The proof can be easily generalized to specify sufficient conditions for a set of macros to be *complete*.

**Lemma 1** *Let $h$ be the heuristic function and $B$ the set of basic operators defined in Table 19. Let $s_g$ be a goal state of $N \times N$ puzzle. Let $s = \langle t_1, \ldots, t_{N^2} \rangle$ be a puzzle state different*

| | |
|---|---|
| Domain parameter | N |
| Initial value | 3 |
| Generate-goal | Let $t_1, \ldots, t_{N^2-1}$ be a random permutation of $1, \ldots, N^2 - 1$. Generate-goal returns $\langle t_1, \ldots, t_{N^2-1}, 0 \rangle$. |
| Basic-operators | $\{u, d, l, r\}$. Let $loc_s(0) = (i_0, j_0)$. $d(s)$ is defined as: $$tile_{u(s)}(i,j) = \begin{cases} tile_s(i_0+1, j_0) & : \ i = i_0, j = j_0 \\ 0 & : \ i = i_0 + 1, j = j_0 \\ tile_s(i,j) & : \ otherwise \end{cases}$$ $d(s)$ is undefined for $i_0 = N - 1$. $u, l, r$ are defined similarly. |
| Heuristic-function | $h(s, s_g) = 4 \cdot N^2 (N^2 - placed(s, s_g)) + 2 \cdot N \cdot d(NextLoc(s), loc_s(NextTile(s))) + d(loc_s(0), loc_s(NextTile(s)))$ |

Table 19: The definitions of the parameters that were used to apply MICRO-HILLARY to the $N \times N$ puzzle domains.

*from $s_g$ such that $solvable(s, s_g)$.*
*If $d\left(loc_s(0), loc_s(nextTile(s))\right) > 1$, then there exists a basic operator $o \in B$ such that $h(o(s), s_g) < h(s, s_g)$.*

**Proof:**

Let $loc_s(NextTile(s)) = \langle i_p, j_p \rangle$, $loc_s(0) = \langle i_0, j_0 \rangle$ and $NextLoc(s) = \langle i_t, j_t \rangle$. The following basic operators will decrease the value of $h$:

$$o = \begin{cases} j_p > j_0 & r \\ j_p = j_0 & \begin{cases} i_p > i_0 & d \\ i_p < i_0 & u \end{cases} \\ j_p < j_0 & \begin{cases} i_p > i_0 & d \\ i_p \leq i_0 & l \end{cases} \end{cases} \tag{4}$$

**Lemma 2** *Let $h$, $B$, $s_g$ and $s$ be defined as in Lemma 1. Let $M$ be the following set of macros:*

$$M = \left\{ \begin{array}{l} lur,\ rul,\ uld,\ ruuld,\ dllur,\ drrul,\ urrdluld,\ urrdluld,\ uuldrdluurd, \\ lurrdluld,\ urdrrullldrrur,\ urdrrullldrrurd,\ llurdrrullldrrurd, \\ uldllurdrulldrrurd \end{array} \right\}$$

*Then the set $M$ is a* complete *set of macros, i.e., there is always an operator $o \in B \cup M$ such that $h(o(s), s_g) < h(s, s_g)$.*

**Proof:**

By Lemma 1, we need to prove only for the cases where $d\left(loc_s(0), loc(NextTile(s))\right) = 1$. There are four possible cases. Tables 20 and 21 show which operator can be applied in each of the cases to decrease the value of $h$. To make the proof simpler, the tables assume $N > 4$.

**Theorem 1** *Let h and M be defined as in Lemma 2.* MICRO-HILLARY*, using M, can solve any solvable $N \times N$ puzzle problem performing no more than $288N^3 - 301N^2$ basic operator applications. The length of the solution is bounded by $50N^3 - 66N^2$.*

**Proof:**

By Lemma 2, we can move from each state to a state with a lower heuristic value. Therefore, after a finite number of states, we will reach the state $s$ with $h(s, s_g) = 0$, which is the goal state. $N^2$ tiles are moved into their goal location. The maximum distance of each tile from its goal location is $2(N-1)$. Each movement of a tile towards its goal location involves moving the blank tile to be adjacent to the next tile, and then using either a basic operator or a macro operator to advance the next tile. At the beginning of this operation the blank tile can be anywhere, so we need to advance it at most $(N-1)+(N-2)$ times to bring it to distance 1 from the next tile. After that, for the remaining $2(N-1)-1$ times, a macro operator can move the blank away from the next tile, but the distance will be at most 4 (this is a property of the specific $M$), so we will need to advance it at most 3 times to bring it next to the next tile. By Lemma 1, the blank tile can be brought to distance 1 using only basic operators. After bringing the empty tile next to the next tile, in the worst case, the algorithm may try all the 4 basic operators and all the macros (with a total length of 124) before finding the one that reduces the heuristic value. All this leads to the following bound on the total number of basic operators applied while solving a problem $p$ using $B$ and $M$:

$$N^2[4((N-1)+(N-2)+3(2(N-1)-1))+2(N-1)(124+4)] = 288N^3 - 301N^2$$

The maximal length of a macro is 18. Therefore, the length of a solution found by MICRO-HILLARY is bounded by

$$N^2[4((N-1)+(N-2)+3(2(N-1)-1))+2(N-1)18] = 50N^3 - 66N^2 \square$$

<table>
<tr><td>$j_p = j_0$</td><td>$i_p > i_0$<br>$(i_p{=}i_0{+}1)$</td><td></td><td></td><td>d</td><td><table><tr><td>1</td><td>2</td><td>3</td><td>4</td><td>5</td></tr><tr><td>6</td><td>7</td><td>8</td><td>10</td><td>18</td></tr><tr><td>20</td><td>19</td><td>17</td><td></td><td>14</td></tr><tr><td>11</td><td>12</td><td>15</td><td>[9]</td><td>16</td></tr><tr><td>23</td><td>22</td><td>13</td><td>24</td><td>21</td></tr></table></td><td><table><tr><td>1</td><td>2</td><td>3</td><td>4</td><td>5</td></tr><tr><td>6</td><td>7</td><td>8</td><td>10</td><td>18</td></tr><tr><td>20</td><td>19</td><td>17</td><td>[9]</td><td>14</td></tr><tr><td>11</td><td>12</td><td>15</td><td></td><td>16</td></tr><tr><td>23</td><td>22</td><td>13</td><td>24</td><td>21</td></tr></table></td></tr>
<tr><td></td><td>$i_p < i_0$<br>$(i_p{=}i_0{-}1)$</td><td>$j_p > j_t$</td><td></td><td>lur</td><td><table><tr><td>1</td><td>2</td><td>3</td><td>4</td><td>5</td></tr><tr><td>6</td><td>16</td><td>23</td><td>14</td><td>10</td></tr><tr><td>15</td><td>17</td><td>22</td><td>12</td><td>20</td></tr><tr><td>11</td><td>13</td><td>24</td><td>[7]</td><td>9</td></tr><tr><td>21</td><td>8</td><td>19</td><td></td><td>18</td></tr></table></td><td><table><tr><td>1</td><td>2</td><td>3</td><td>4</td><td>5</td></tr><tr><td>6</td><td>16</td><td>23</td><td>14</td><td>10</td></tr><tr><td>15</td><td>17</td><td>22</td><td>12</td><td>20</td></tr><tr><td>11</td><td>13</td><td>[7]</td><td></td><td>9</td></tr><tr><td>21</td><td>8</td><td>24</td><td>19</td><td>18</td></tr></table></td></tr>
<tr><td></td><td></td><td>$j_p < j_t$</td><td></td><td>rul</td><td><table><tr><td>1</td><td>2</td><td>3</td><td>4</td><td>5</td></tr><tr><td>6</td><td>7</td><td>8</td><td>15</td><td>12</td></tr><tr><td>18</td><td>[9]</td><td>10</td><td>17</td><td>20</td></tr><tr><td>13</td><td></td><td>22</td><td>11</td><td>14</td></tr><tr><td>19</td><td>21</td><td>16</td><td>24</td><td>23</td></tr></table></td><td><table><tr><td>1</td><td>2</td><td>3</td><td>4</td><td>5</td></tr><tr><td>6</td><td>7</td><td>8</td><td>15</td><td>12</td></tr><tr><td>18</td><td></td><td>[9]</td><td>17</td><td>20</td></tr><tr><td>13</td><td>22</td><td>10</td><td>11</td><td>14</td></tr><tr><td>19</td><td>21</td><td>16</td><td>24</td><td>23</td></tr></table></td></tr>
<tr><td></td><td></td><td>$j_p = j_t$</td><td>$j_p < N$</td><td>ruuld</td><td><table><tr><td>1</td><td>2</td><td>3</td><td>4</td><td>5</td></tr><tr><td>6</td><td>7</td><td>8</td><td>9</td><td>10</td></tr><tr><td>11</td><td>12</td><td>13</td><td>20</td><td>19</td></tr><tr><td>21</td><td>22</td><td>16</td><td>[14]</td><td>18</td></tr><tr><td>24</td><td>17</td><td>23</td><td></td><td>15</td></tr></table></td><td><table><tr><td>1</td><td>2</td><td>3</td><td>4</td><td>5</td></tr><tr><td>6</td><td>7</td><td>8</td><td>9</td><td>10</td></tr><tr><td>11</td><td>12</td><td>13</td><td>[14]</td><td>20</td></tr><tr><td>21</td><td>22</td><td>16</td><td></td><td>19</td></tr><tr><td>24</td><td>17</td><td>23</td><td>15</td><td>18</td></tr></table></td></tr>
<tr><td></td><td></td><td></td><td>$j_p = N$</td><td>uuldrdluurd</td><td><table><tr><td>1</td><td>2</td><td>3</td><td>4</td><td>5</td></tr><tr><td>6</td><td>7</td><td>8</td><td>9</td><td>13</td></tr><tr><td>21</td><td>20</td><td>19</td><td>14</td><td>[10]</td></tr><tr><td>16</td><td>22</td><td>23</td><td>18</td><td></td></tr><tr><td>24</td><td>17</td><td>12</td><td>11</td><td>15</td></tr></table></td><td><table><tr><td>1</td><td>2</td><td>3</td><td>4</td><td>5</td></tr><tr><td>6</td><td>7</td><td>8</td><td>9</td><td>[10]</td></tr><tr><td>21</td><td>20</td><td>19</td><td>14</td><td></td></tr><tr><td>16</td><td>22</td><td>23</td><td>13</td><td>18</td></tr><tr><td>24</td><td>17</td><td>12</td><td>11</td><td>15</td></tr></table></td></tr>
</table>

Table 20: The macro used for each of the conditions on the tile indices. The right column shows an example for a state before and after the macro application.

| | | | | | | |
|---|---|---|---|---|---|---|
| $j_p > j_0$<br>($j_p$=$j_0$+1) | $i_p = i_0$ | $j_p > j_t$ | | r | 1 2 3 4 5<br>6 7 8 9 10<br>11 22  [12] 17<br>18 13 15 14 20<br>19 21 16 24 23 | 1 2 3 4 5<br>6 7 8 9 10<br>11 22 [12]  17<br>18 13 15 14 20<br>19 21 16 24 23 |
| | | $j_p < j_t$ | $i_p < N$ | drrul | 1 2 3 4 5<br>6 7 8 9 12<br>18 11  [10] 20<br>13 15 17 22 14<br>19 21 16 24 23 | 1 2 3 4 5<br>6 7 8 9 12<br>18 11 17  [10]<br>13 15 22 14 20<br>19 21 16 24 23 |
| | | | $i_p = N$<br>$i_t < N-1$ | urrdluld | 1 2 3 4 5<br>6 7 8 9 10<br>11 12 13 14 18<br>23 20 16 17 22<br> [15] 21 19 24 | 1 2 3 4 5<br>6 7 8 9 10<br>11 12 13 14 18<br>23 20 21 17 22<br> 16 [15] 19 24 |
| | | | $i_p = N$<br>$i_t = N-1$<br>$j_t - j_p \leq 1$ | urrdluld | 1 2 3 4 5<br>6 7 8 9 10<br>11 12 13 14 15<br>16 17 18 22 20<br>21  [19] 24 23 | 1 2 3 4 5<br>6 7 8 9 10<br>11 12 13 14 15<br>16 17 18 24 20<br>21  22 [19] 23 |
| | | | $i_p = N$<br>$i_t = N-1$<br>$j_t - j_p > 1$ | urdrrullldrrurd | 1 2 3 4 5<br>6 7 8 9 10<br>11 12 13 14 15<br>16 17 18 19 22<br>21  [20] 24 23 | 1 2 3 4 5<br>6 7 8 9 10<br>11 12 13 14 15<br>16 17 18 19 22<br>21 24 23 [20]  |
| | | $j_p = j_t$ | $j_0 > 1$ | lurrdluld | 1 2 3 4 5<br>6 7 8 9 12<br>18 11 17  [10]<br>13 15 22 14 20<br>19 21 16 24 23 | 1 2 3 4 5<br>6 7 8 9 [10]<br>18 11  12 17<br>13 15 22 14 20<br>19 21 16 24 23 |
| | | | $j_0 = 1$ | urdrulldrur | 1 2 3 4 5<br>6 7 8 9 10<br>11 12 13 14 15<br>16 23 20 18 19<br> [17] 21 24 22 | 1 2 3 4 5<br>6 7 8 9 10<br>11 12 13 14 15<br>16 [17]  18 19<br>21 23 20 24 22 |
| $j_p < j_0$<br>($j_p$=$j_0$−1) | $i_p = i_0$ | $j_p > j_t$ | $i_t < N$<br>$i_p < N$ | dllur | 1 2 3 4 5<br>6 7 8 9 10<br>11 12 13 14 15<br>16 21 [17]  20<br>18 19 24 22 23 | 1 2 3 4 5<br>6 7 8 9 10<br>11 12 13 14 15<br>16 [17]  22 20<br>18 21 19 24 23 |
| | | | $i_t < N$<br>$i_p = N$ | uld | 1 2 3 4 5<br>6 7 8 9 10<br>11 12 13 14 15<br>16 17 22 21 23<br>20 19 24 [18]  | 1 2 3 4 5<br>6 7 8 9 10<br>11 12 13 14 15<br>16 17 22 [18] 21<br>20 19 24  23 |
| | | | $i_t = N$<br>$j_0 < N$ | llurdrrullldrrurd | 1 2 3 4 5<br>6 7 8 9 10<br>11 12 13 14 15<br>16 17 18 19 20<br>22 23 [21]  24 | 1 2 3 4 5<br>6 7 8 9 10<br>11 12 13 14 15<br>16 17 18 19 20<br>22 [21] 24 23  |
| | | | $i_t = N$<br>$j_0 = N$<br>$J_t < N-2$ | uldllurdrulldrrurd | 1 2 3 4 5<br>6 7 8 9 10<br>11 12 13 14 15<br>16 17 18 19 20<br>22 24 23 [21]  | 1 2 3 4 5<br>6 7 8 9 10<br>11 12 13 14 15<br>16 17 18 19 20<br>22 23 [21] 24  |
| | | | $i_t = N$<br>$j_0 = N$<br>$J_t \geq N-2$ | Unsolvable puzzle | 1 2 3 4 5<br>6 7 8 9 10<br>11 12 13 14 15<br>16 17 18 19 20<br>21 22 24 [23]  | |
| | | $j_p < j_t$ | | l | 1 2 3 4 5<br>6 7 8 9 10<br>11 12 13 14 15<br>16 17 19 22 20<br>[18]  21 24 23 | 1 2 3 4 5<br>6 7 8 9 10<br>11 12 13 14 15<br>16 17 19 22 20<br> [18] 21 24 23 |
| | | | $j_p = j_t$ | uld | 1 2 3 4 5<br>6 7 8 9 10<br>11 20 17 16 18<br>19 [12]  15 14<br>23 22 13 24 21 | 1 2 3 4 5<br>6 7 8 9 10<br>11 [12] 20 16 18<br>19  17 15 14<br>23 22 13 24 21 |

Table 21: (Cont.) The macro used for each of the conditions on the tile indices. The right column shows an example for a state before and after the macro application.

We can use the same lemmas to prove that MICRO-HILLARY can solve any solvable problem in $O(N^3)$ even without learning simply by performing BFS search in each place where a macro is used in the original proof. However, instead of using 128 in the above formula, we will use $4^{18}$ (18, the length of the longest macro, can be used as an upper bound on the depth of the search). This gives a bound of $137,438,953,504N^3 - 137,438,953,520N^2$, which is about 477,218,588 times larger than the constant for MICRO-HILLARY after learning. While in the pure sense of computational complexity both bounds belong to the same complexity class, the huge constant makes this fact meaningless for any practical purpose.

## References

Bui, T. (1997). Solving the 8-puzzle with genetic programming. In Koza, J. R. (Ed.), *Genetic Algorithms and Genetic Programming at Stanford*, pp. 11–17. Stanford Bookstore, Stanford, California.

Cohen, W. W. (1992). Using distribution-free learning theory to analyze solution path caching mechanisms. *Computational Intelligence*, *8*(2), 336–375.

Etzioni, O., & Etzioni, R. (1994). Statistical methods for analyzing speedup learning experiments. *Machine Learning*, *14*, 333–347.

Fikes, R. E., Hart, P. E., & Nilsson, N. J. (1972). Learning and executing generalized robot plans. *Artificial Intelligence*, *3*, 251–288.

Finkelshtein, L., & Markovitch, S. (1992). Selective acquisition and utilization of macro operators: A learning program solves the general N X N puzzle. Technical report CIS9216, Computer Science Department, Technion, Haifa, Israel.

Ginsberg, M. L., & Harvey, W. D. (1992). Iterative broadening. *Artificial Intelligence*, *55*(2-3), 367–383.

Gratch, J., & DeJong, D. (1992). COMPOSER: A probabilistic solution to the utility problem in speed-up learning. In *Proceedings of the Tenth National Conference on Artificial Intelligence*, pp. 235–240. San Jose, California. Morgan Kaufmann.

Greiner, R., & Likuski, J. (1989). Incorporating redundant learned rules: A preliminary formal analysis of EBL. In *Proceedings of the Eleventh International Joint Conference on Artificial Intelligence*, pp. 744–749. Detroit, Michigan. Morgan Kaufmann.

Iba, G. A. (1985). Learning by discovering macros in puzzle solving. In *Proceedings of the Ninth International Joint Conference on Artificial Intelligence*, pp. 640–642. Los Angeles, California. Morgan Kaufmann.

Iba, G. A. (1989). A heuristic approach to the discovery of macro-operators. *Machine Learning*, *3*(4), 285–317.

Iba, W., Wogulis, J., & Langley, P. (1988). Trading off simplicity and coverage in incremental concept learning. In *Proceedings of the Fifth International Conference on Machine Learning*, pp. 73–79. Ann Arbor, MI. Morgan Kaufmann.

Johnson, W. W., & Story, W. E. (1879). Notes on the "15" puzzle. *American Journal of Mathematics*, *2*, 397–404.

Korf, R. E. (1985). *Learning to Solve Problems by Searching for Macro-Operators.* Pitman, Boston.

Korf, R. E. (1993). Linear-space best-first search. *Artificial Intelligence*, *62*(1), 41–78.

Korf, R. E. (1985). Depth-first iterative-deepening: An optimal admissible tree search. *Artificial Intelligence*, *27*(1), 97–109.

Korf, R. E. (1990). Real-time search for dynamic planning. In *Proceedings of the AAAI Spring Symposium on Planning in Uncertain, Unpredictable, or Changing Environments*, pp. 72–76. Stanford, CA.

Korf, R. E., & Taylor, L. A. (1996). Finding optimal solutions to the twenty-four puzzle. In *Proceedings of the Fourteenth National Conference on Artificial Intelligence*, pp. 1202–1207. Portland, Oregon. Morgan Kaufmann.

Kraitchik, M. (1953). *Mathematical Recreations.* Dover, New York.

Laird, J. E., Rosenbloom, P. S., & Newell, A. (1986). Chunking in soar: The anatomy of a general learning mechanism. *Machine Learning*, *1*, 11–46.

Markovitch, S., & Rosdeutscher, I. (1992). Systematic experimentation with deductive learning: Satisficing vs. optimizing search. In *Proceedings of the Knowledge Compilation and Speedup Learning Workshop.* Aberdeen, Scotland.

Markovitch, S., & Scott, P. D. (1988). The role of forgetting in learning. In *Proceedings of The Fifth International Conference on Machine Learning*, pp. 459–465. Ann Arbor, MI. Morgan Kaufmann.

Markovitch, S., & Scott, P. D. (1993). Information filtering: Selection mechanisms in learning systems. *Machine Learning*, *10*(2), 113–151.

Minton, S. (1985). Selectively generalizing plans for problem solving. In *Proceedings of the Ninth International Joint Conference on Artificial Intelligence*, pp. 596–599. Los Angeles, CA. Morgan Kaufmann.

Minton, S. (1988). *Learning Search Control Knowledge: An Explanation-Based Approach.* Kluwer, Boston, MA.

Mitchell, T. M., Utgoff, P. E., & Banerji, R. (1983). Learning by experimentation: Acquiring and refining problem-solving heuristics. In Michalski, R. S., Carbonell, J. G., & Mitchell, T. M. (Eds.), *Machine Learning: An Artificial Intelligence Approach.* Tioga, Palo Alto, California.

Mooney, R. (1989). The effect of rule use on the utility of explanation-based learning. In *Proceedings of the Eleventh International Joint Conference of Artificial Intelligence*, pp. 725–730. Detroit, Michigan. Morgan Kaufmann.

Natarajan, B. K. (1989). On learning from exercises. In *Proc. 2nd Annu. Workshop on Comput. Learning Theory*, pp. 72–87 San Mateo, CA. Morgan Kaufmann.

Nilsson, N. J. (1980). *Principles of Artificial Intelligence.* Tioga, Palo Alto, CA.

Pearl, J. (1984). *Heuristics: intelligent search strategies for computer problem solving.* Addison-Wesley, Reading, Massachusetts.

Ratner, D., & Warmuth, M. (1986). Finding a shortest solution for the $n \times n$ extension of the 15-puzzle is intractable. In *Proceedings of the Fifth National Conference on Artificial Intelligence (AAAI-86)*, Vol. 1, pp. 168–172. Philadelphia, Pennsylvania. Morgan Kaufmann.

Ratner, D., & Warmuth, M. (1990). The $(N^2 - 1)$-puzzle and related relocation problems. *Journal of Symbolic Computation*, *10*, 111–137.

Reddy, C., & Tadepalli, P. (1997). Learning goal-decomposition rules using exercises. In *Proceedings of the Fourteenth International Conference on Machine Learning.* Nashville, TN. Morgan Kaufmann.

Ruby, D., & Kibler, D. (1989). Learning subgoal sequences for planning. In *Proceedings of the Eleventh International Joint Conference on Artificial Intelligence*, pp. 609–614. Detroit, Michigan. Morgan Kaufmann.

Ruby, D., & Kibler, D. (1992). EASe: Integrating search with learning episodes. Technical report 92-30, Information and Computer Science, University of California, Irvine, California.

Schofield, P. D. A. (1967). Complete solution of the eight-puzzle. In Collins, N. L., & Michie, D. (Eds.), *Machine Intelligence*, Vol. 1. Oliver and Boyd, Edinburgh.

Segre, A., Elkan, C., & Russell, A. (1991). A critical look at experimental evaluation of EBL. *Machine Learning*, *6*, 183–195.

Shavlik, J. W. (1990). Acquiring recursive and iterative concepts with explanation-based learning. *Machine Learning*, *5*, 39–70.

Simon, H. A., & Kadane, J. B. (1975). Optimal problem-solving search: All-or-none solution. *Artificial Intelligence*, *6*, 235–247.

Subramanian, D., & Hunter, S. (1992). Measuring utility and the design of provably good EBL algorithms. In *Proceedings of the Ninth International Workshop on Machine Learning*, pp. 426–435. Aberdeen, Scotland. Morgan Kaufmann.

Tadepalli, P., & Natarajan, B. K. (1996). A formal framework for speedup learning from problems and solutions. *Journal of Artificial Intelligence Research*, *4*, 419–443.

Tadepalli, P. (1991). A formalization of explanation-based macro-operator learning. In *Proceedings of International Joint Conference for Artificial Intelligence*, pp. 616–622. Sydney, Australia. Morgan Kaufmann.

Tait, P. G. (1880). Note on the theory of the "15 puzzle". *Proceedings of the Royal Society of Edinburgh*, *10*, 664–665.